\documentclass[pdflatex,sn-mathphys-num,iicol]{sn-jnl}% Math and Physical Sciences Numbered Reference Style
\usepackage{graphicx}%
\usepackage{multirow}%
\usepackage{amsmath,amssymb,amsfonts}%
\usepackage{amsthm}%
\usepackage{mathrsfs}%
\usepackage[title]{appendix}%
\usepackage{textcomp}%
\usepackage{manyfoot}%
\usepackage{booktabs}%
\usepackage{algorithm}%
\usepackage{algorithmicx}%
\usepackage{algpseudocode}%
\usepackage{listings}%
\usepackage{tabularx}%
\usepackage{lscape}%
\usepackage{comment}%
\usepackage{rotating}%
\usepackage[table]{xcolor}%
\usepackage{pgfplots}%
\usepackage{pgfplotstable}%
\usepackage{adjustbox}%
\pgfplotsset{compat=1.18}%
\usepackage{booktabs}%
\usepackage{tikz}%
\usepackage{fix-cm}
\theoremstyle{thmstyleone}%
\theoremstyle{thmstyletwo}%

\theoremstyle{thmstylethree}%

\begin{document}

\title[Article Title]{Hallucinations in LLMs: A Lifecycle-Based Survey of Causes, Detection, Mitigation, and Prevention}

%%=============================================================%%
%% Author Details
%%=============================================================%%

\author*[1]{\fnm{Naveen} \sur{Lamba}}\email{naveenlamba30894@gmail.com}

\author[1]{\fnm{Sanju} \sur{Tiwari}}\email{tiwarisanju18@ieee.org}

\author[2]{\fnm{Manas} \sur{Gaur}}\email{manas@umbc.edu}

%%=============================================================%%
%% Affiliation
%%=============================================================%%

\affil[1]{\orgdiv{CAIMIF \& Department of CSA},
\orgname{Sharda University},
\city{Greater Noida},
\country{India}}

\affil[2]{\orgname{University of Maryland, Baltimore County},
\city{Baltimore},
\country{USA}}


\abstract{The lifecycle of hallucination in LLMs is a concept that enables building solid frameworks on the control and reliability of LLMs in high-stakes environments, including health, legal, and scientific research. Although previous surveys have primarily focused on detection or mitigation, this survey provides a lifecycle-based overview of the hallucinations in the LLMs, their cause, detection, mitigation, and prevention.We propose a three-fold categorization of hallucinations across the LLM lifecycle: data-related, training-related, and inference-related, which is consistent with the lifecycle of the development of the LLM. Each of these stages is discussed regarding the cause of hallucinations, their detection, and the ways they can be addressed under specific mitigation or prevention interventions. In addition, we discuss the available benchmark data using a number of parameters so as to establish their suitability in identifying, restricting and managing hallucinations. The survey provides researchers and practitioners with a standardized framework to understand, diagnose, and cure hallucinations in a systematic system to present actionable data to build safer and more reliable LLMs.}




\maketitle

\section{Introduction}\label{sec:Introduction}

Large language models (LLMs) are becoming increasingly popular across domains such as law, scientific research, and healthcare due to their strong generative and reasoning capabilities \cite{10850911, zheng-etal-2025-automation, Maity2025HealthcareLLM}. Recent studies have explored their applications in legal reasoning, scientific discovery, and clinical decision support systems \cite{enguehard-etal-2025-lemaj, Zhang2025ScientificMethod, Pal2023}. LLMs are typically trained on large scale corpora, and then fine tuned to comply with user instructions.  Prominent examples include Llama 2 \cite{Touvron2023l}, Claude \cite{Anthropic2023}, and GPT-4 \cite{OpenAI2023}. Other recent LLMs have also been endowed with reasoning capabilities to solve intricate cognitive problems like mathematical problem solving, multi-step logical reasoning, code synthesis and decision making in ambiguous situations \cite{DeepSeek-AI2024, Grattafiori2024}. In spite of these developments, there are still challenges especially in the real time assessment and the control of hallucinations. 

Hallucination refers to the phenomenon where LLMs produce content that is factually incorrect, logically contradictory, or fabricated without a reliable grounding basis \cite{Ji2023, Peng2023}. Hallucinations in high-stakes areas can be life threatening, such as misinformation in health care, misunderstandings in legal contexts and incorrect scientific findings \cite{Lin2022, OpenAI2023}. It is thus important to comprehend the causes of hallucinations and to manage them systematically in order to deploy it safely.

\begin{figure*}[!t]
    \centering
    \includegraphics[width=1\linewidth]{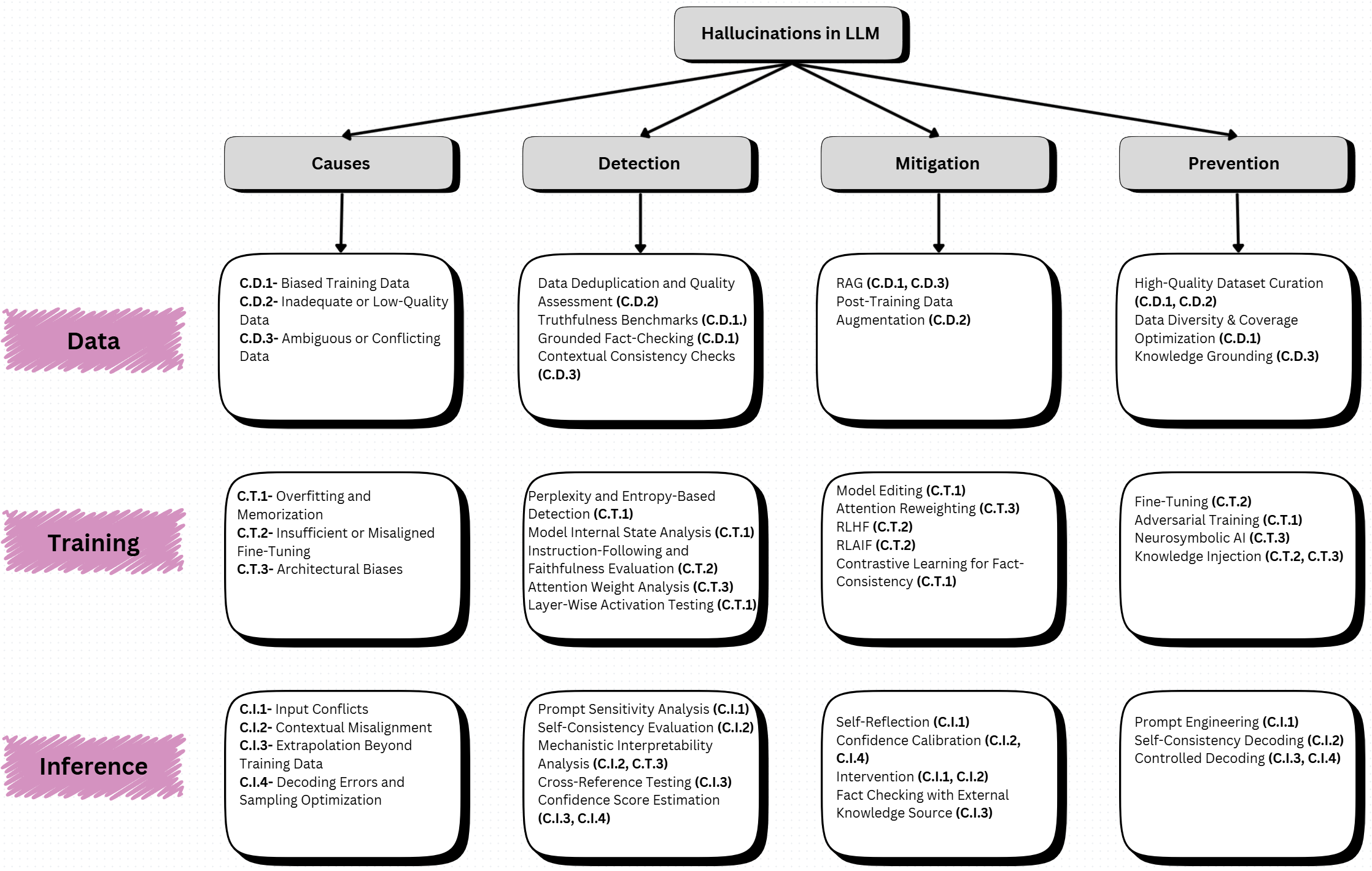}
    \caption{Lifecycle-based framework of hallucinations in LLMs}
    \label{fig:lifecycle}
\end{figure*}

\textbf{Current Survey Perspective.} We take a \textit{lifecycle perspective} of hallucinations in LLMs in this survey. We define the \textbf{hallucination lifecycle} as the set of stages across the LLM pipeline in which hallucinations may emerge, be detected, mitigated, or proactively prevented: \texttt{Cause}, \texttt{Detection}, \texttt{Mitigation}, and \texttt{Prevention}. \autoref{fig:lifecycle} shows a Lifecycle-based framework of hallucinations in LLMs across four stages, Cause, Detection, Mitigation, and Prevention, mapped onto the LLM pipeline phases: Data, Training, and Inference. It systematically links hallucination causes to their corresponding detection, mitigation, and prevention techniques. Each method is annotated with the specific cause(s) it addresses: data-related (C.D.1–C.D.3), training-related (C.T.1–C.T.3), and inference-related (C.I.1–C.I.4). All lifecycle phases are further studied in relation to the three stages of the LLM pipeline data, training and inference.

\begin{itemize}
    \item \texttt{Cause} examines the role played by data quality, training dynamics and inference behavior in hallucinations.
    \item \texttt{Detection} looks into ways of spotting hallucination through the consideration of coherence, factual accuracy, and matchability of the context.
    \item \texttt{Mitigation} is aimed at minimizing the effect of the hallucinations on outputs after they have taken place.
    \item \texttt{Prevention} deals with upstream measures of constraining the incidence of hallucinations like quality of the information and sound training guidelines.
\end{itemize}

In order to transform the lifecycle perspective into a synthetic rather than simply a descriptive overview of the literature, the survey will be informed by specific research questions that organize the selection, categorization, and interpretation of the literature. In particular, developing the overview around specific questions helps assure that the lifecycle model will be usable not only conceptually but also analytically.

The research questions addressed by this survey are:

\begin{itemize}
    \item \texttt{RQ1:} What is hallucination in LLMs and how can it be understood within a lifecycle framework?
    \item \texttt{RQ2:} What are the causes of hallucinations across data, training, and inference stages?
    \item \texttt{RQ3:} What techniques exist for detecting hallucinations?
    \item \texttt{RQ4:} What mitigation and prevention strategies reduce hallucination risks?
    \item \texttt{RQ5:} How suitable are existing benchmarks for evaluating hallucinations?
\end{itemize}

\begin{table*}[!t]
\centering
\caption{Comparison of existing hallucination surveys across different phases of the hallucination lifecycle. Train = Training; Infer = Inference }
\label{tab:comparison}

\setlength{\tabcolsep}{1.5pt}

\begin{tabular}{l c c c c c c c c c c c c c c}
\toprule
 & \multicolumn{12}{c}{\textbf{Hallucination lifecycle}} &  &  \\
 
 \cmidrule(lr){2-13}
 
 & \multicolumn{3}{c}{\textbf{Causes}} & \multicolumn{3}{c}{\textbf{Detection}} & \multicolumn{3}{c}{\textbf{Mitigation}} & \multicolumn{3}{c}{\textbf{Prevention}} & & \\
 
 \cmidrule(lr){2-13}

\textbf{Survey}
& \rotatebox{90}{Data}
& \rotatebox{90}{Train}
& \rotatebox{90}{Infer}
& \rotatebox{90}{Data}
& \rotatebox{90}{Train}
& \rotatebox{90}{Infer}
& \rotatebox{90}{Data}
& \rotatebox{90}{Train}
& \rotatebox{90}{Infer}
& \rotatebox{90}{Data}
& \rotatebox{90}{Train}
& \rotatebox{90}{Infer}
& \shortstack{\textbf{Suitable}\\\textbf{Benchmark}}
& \shortstack{\textbf{No. of}\\\textbf{Aspects}} \\

\midrule
\citet{Rawte2023} & $\times$ & $\times$ & $\times$ & $\times$ & $\times$ & $\times$ & $\times$ & $\times$ & $\times$ & $\times$ & $\times$ & $\times$ & $\times$ & 0 \\
\citet{Zhang2023} & \checkmark & \checkmark & \checkmark & $\times$ & $\times$ & $\times$ & \checkmark & \checkmark & \checkmark & $\times$ & $\times$ & $\times$ & $\times$ & 6 \\
\citet{Ji2023} & \checkmark & \checkmark & \checkmark & $\times$ & $\times$ & $\times$ & \checkmark & \checkmark & \checkmark & $\times$ & $\times$ & $\times$ & $\times$ & 6 \\
\citet{Ye2023} & $\times$ & $\times$ & $\times$ & $\times$ & $\times$ & $\times$ & \checkmark & \checkmark & \checkmark & $\times$ & $\times$ & $\times$ & $\times$ & 3 \\
\citet{Huang2024} & \checkmark & \checkmark & \checkmark & $\times$ & $\times$ & $\times$ & \checkmark & \checkmark & \checkmark & $\times$ & $\times$ & $\times$ & $\times$ & 6 \\
\citet{Andriopoulos2023} & \checkmark & $\times$ & $\times$ & $\times$ & $\times$ & $\times$ & \checkmark & $\times$ & \checkmark & \checkmark & $\times$ & $\times$ & $\times$ & 4 \\
\citet{Bai2025} & \checkmark & \checkmark & \checkmark & $\times$ & $\times$ & $\times$ & \checkmark & \checkmark & \checkmark & $\times$ & $\times$ & $\times$ & $\times$ & 6 \\
\citet{Lavrinovics2024} & $\times$ & $\times$ & $\times$ & $\times$ & $\times$ & $\times$ & \checkmark & $\times$ & \checkmark & $\times$ & $\times$ & $\times$ & $\times$ & 2 \\
\citet{Tonmoy2024} & $\times$ & $\times$ & $\times$ & $\times$ & $\times$ & $\times$ & \checkmark & \checkmark & \checkmark & \checkmark & \checkmark & \checkmark & $\times$ & 6 \\
\citet{Agrawal2023} & $\times$ & $\times$ & $\times$ & $\times$ & $\times$ & $\times$ & \checkmark & $\times$ & \checkmark & \checkmark & \checkmark & $\times$ & $\times$ & 4 \\

\midrule
\rowcolor{gray!10}\textbf{Our Survey} & \checkmark & \checkmark & \checkmark & \checkmark & \checkmark & \checkmark & \checkmark & \checkmark & \checkmark & \checkmark & \checkmark & \checkmark & \checkmark & 13 \\
\bottomrule
\end{tabular}
\end{table*}

All major sections of the survey are structured around answering one or more of these questions. The final discussion section is an exception, as it reviews the research questions to summarize relevant findings and identify gaps that are being left for further research.

\textbf{Connection with Existing Surveys.} \autoref{tab:comparison} compares our survey with prior work. It also clarifies how we define the 13 aspects: the four lifecycle stages, \texttt{Cause}, \texttt{Detection}, \texttt{Mitigation}, and \texttt{Prevention}, each analyzed across the three LLM pipeline phases (\textit{data}, \textit{training}, and \textit{inference}), together with a dedicated dimension for \texttt{benchmarking}. This yields 13 aspects in total. 
Most existing surveys cover only one or two stages of the hallucination lifecycle, typically focusing on mitigation and prevention. Consequently, existing findings are often fragmented and do not systematically associate hallucinations with specific phases of the LLM lifecycle \cite{Peng2023}. Compared to mitigation-focused surveys such as \citet{Tonmoy2024} and \citet{Agrawal2023}, the proposed framework additionally incorporates data- and training-stage factors that influence hallucination behavior. This broader perspective highlights how vulnerabilities introduced earlier in the LLM pipeline may contribute to downstream generation errors and therefore motivates lifecycle-wide intervention strategies.
We additionally discuss interpretive parallels between hallucination-related behaviors and cognitive phenomena such as source amnesia, confabulation, and suggestibility \cite{Ellis2020, McClelland1995, Schacter2021}. These cognitive parallels are intended primarily as heuristic and explanatory analogies rather than formal cognitive equivalence claims between human cognition and LLM behavior.

\textbf{Benchmark Analysis.} Current hallucination measuring benchmarks frequently focus on single tasks, such as question answering, summarization, or dialogue \cite{Lin2022,Heo2025,Min2023}, without the development and growth of hallucinations through the LLM lifecycle. These datasets are usually not based on verifiable external knowledge, have poor interpretability, and cannot reflect dynamic inference behavior. To overcome this, we suggest an overall evaluation framework (\autoref{tab:suitability}) with 15 criteria to analyze the appropriateness of benchmark to various lifecycle phases. The framework identifies the shortcomings in the existing data collections and underscores the necessity of multi-dimensional benchmarks to facilitate layer-wise diagnoses, chain-of-thought verification and domain-specific hallucination monitoring \cite{Peng2023, Ji2023}. From a lifecycle perspective, benchmark suitability depends not only on task-level accuracy measurement, but also on the ability to analyze how hallucinations emerge, propagate, and are controlled across different stages of the LLM pipeline.

\textbf{Key Contributions.} The contributions of this survey are as follows:

\begin{itemize}
    \item We propose a lifecycle-based framework that connects hallucination causes, detection signals, mitigation strategies, and prevention techniques across data, training, and inference stages.
    \item We synthesize hallucination causes using a Root–Trigger–Manifestation framework that explains how hallucinations propagate through the LLM pipeline.
    \item We organize hallucination detection techniques based on observable indicators derived from lifecycle instability signals.
    \item We present a comparative analysis of mitigation and prevention strategies, including quantitative comparisons of effectiveness and computational cost.
    \item We evaluate existing hallucination benchmarks and identify limitations in measuring hallucination behavior across lifecycle stages.
\end{itemize}

\textbf{Survey Organization.} The rest of this survey is organized according to the proposed lifecycle framework. \autoref{sec:cause} will discuss the causes of hallucinations in data, training, and inference; \autoref{sec:Detection} will give an overview of Detection techniques in the same fashion, \autoref{sec:mitigation} will address mitigation techniques, and \autoref{sec:prevention} will address proactive prevention techniques. Lastly in \autoref{sec:benchmark} it would appraise the current benchmarks on a multi criteria matrix to evaluate how suitable it was to handle hallucinations during the lifetime of the LLM.

\section{Methodology}

\begin{figure}[ht]
    \centering
    \includegraphics[width=1\linewidth]{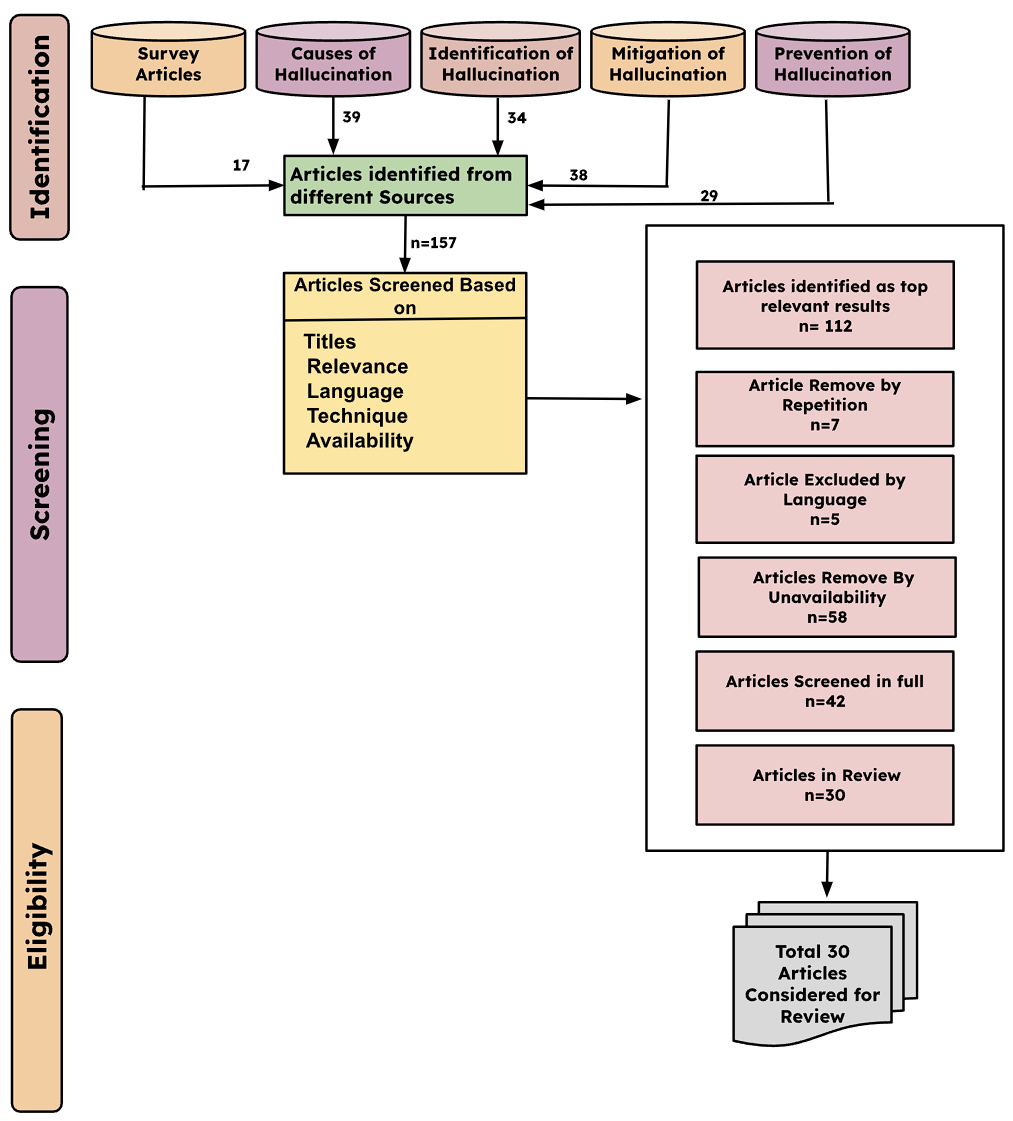}
    \caption{PRISMA style flow diagram illustrating the article selection process for the survey}
    \label{fig:prisma}
\end{figure}

The survey follows a structured and transparent evaluation methodology designed to ensure consistency, traceability, and reproducibility throughout the review process. Rather than relying solely on narrative synthesis, the methodology incorporates systematic literature identification, screening, eligibility assessment, lifecycle-based coding, and comparative analysis. This process establishes a traceable relationship between the selected literature and the analytical findings presented throughout the survey. A PRISMA-style flow diagram summarizing the article selection and evaluation process is shown in \autoref{fig:prisma}.

\subsection{Literature Identification}

Various works were collected from different academic resources to vary between peer-reviewed articles and important technical preprints in this constantly evolving field. IEEE Xplore, ACM Digital Library, arXiv, and Google Scholar were also consulted to ensure balance between solid and lasting material and recent works pertaining to LLMs.

The search focused on keyword combinations related to hallucination and factual inconsistency in LLMs. Representative query terms included: \emph{``LLM hallucination'', ``hallucination in large language models'', ``factual inconsistency in generative models'', ``hallucinated text generation'', and ``reliability of large language models''.}

The search for the articles included publications between 2020 and 2025, the period when instruction-tuned LLMs started to gain wider attention. The initial search retrieved 157 candidate articles. These articles include survey articles, causal analysis, detection techniques, mitigation strategies, and benchmark studies. To ensure that the search did not miss other essential articles, we also referred to the reference list of highly cited articles.

\subsection{Inclusion and Exclusion Criteria}

Explicit inclusion and exclusion criteria were applied to maintain consistency across screening.

\noindent Studies were included if they:

\begin{itemize}
    \item Directly addressed hallucination, factual inconsistency, or reliability in LLMs
    \item Provided empirical, methodological, or benchmark-based analysis
    \item Were written in English
    \item Were peer-reviewed or widely recognized technical preprints
    \item Contained sufficient methodological detail to support lifecycle classification
\end{itemize}

\noindent Studies were excluded if they:

\begin{itemize}
    \item Consisted primarily of opinion or editorial commentary
    \item Were unrelated to hallucination in LLMs
    \item Duplicated previously identified records
    \item Had inaccessible or incomplete full text
    \item Lacked substantive technical contribution
\end{itemize}

These criteria were designed to ensure that the final corpus reflects substantive research rather than anecdotal discussion. The exclusion process was designed to prioritize studies with sufficient technical depth and direct relevance to lifecycle-oriented hallucination analysis. Papers that only mentioned hallucination phenomena tangentially, lacked methodological transparency, or did not provide analyzable technical contributions were excluded to maintain consistency and comparability across the reviewed literature.

\subsection{Screening and Eligibility Assessment}

The screening consisted of two steps. Firstly, titles and/or abstracts were used as search tools to identify credible and related literature on the subject. From 157 articles, 112 articles made it through the first round of filtering based on relevance criteria. Then came the eligibility checks. Seven did not make it through as they were considered duplicates. Five studies were excluded due to language constraints, and 58 were disqualified because full-text versions were unavailable. Then 42 underwent further screening as full text and were checked for their relevance and appropriateness in respect to the lifecycle framework criteria. Finally, 30 made it through to the end. The final analytical corpus was therefore intended to represent a focused lifecycle-oriented synthesis of methodologically relevant hallucination research rather than an exhaustive catalog of all LLM-related publications.

\subsection{Data Extraction}

Each study within the review is analyzed using a standardized process of extracting data, making it possible to compare the results of different approaches. The aim of extracting data is not to list all approaches but to obtain some broad attributes that clarify why a study is a part of the lifecycle of hallucinations.

The study selection process emphasized representative and methodologically influential works corresponding to major hallucination-related techniques and lifecycle stages. Rather than maximizing the total number of included papers, the survey prioritized technically significant and analytically relevant studies that could serve as representative exemplars for lifecycle-oriented comparison and synthesis. Although the synthesis tables often highlight a single representative study for concise lifecycle classification, the accompanying discussions incorporate multiple related works associated with each technique category.

For each of the papers, we identified a number of analytical dimensions, including the lifecycle stage the work addressed (cause, detection, mitigation, or prevention); the corresponding phase in the associated machine learning pipeline (data, training, or inference); the type of intervention or analysis employed; the criterion for evaluating the work; and the strength of the associated evidence base. Here, a number of diverse studies can be compared on the basis of a common factor. Such an abstract construct prevents it from being locked into a specific set of methods and keeps it expandable with new techniques that arise. A consistent extraction template was applied across all included studies to support systematic lifecycle-oriented comparison and synthesis.

\subsection{Lifecycle Coding Rules}

To make the lifecycle taxonomy work, we have applied clear and explicit decision rules in classifying each of the studies. A study could be marked as:

\begin{itemize}
    \item \textbf{Cause:} If it explains the mechanisms or conditions that produce hallucination
    \item \textbf{Detection:} If it detects hallucination after it occurs
    \item \textbf{Mitigation:} If it improves or trims a post-generated hallucinated output
    \item \textbf{Prevention:} If it changes upstream data and/or training processes in some manner to decrease hallucinations
\end{itemize}

When a study affected more than one stage, its classification included each relevant category. We were able to do this by indicating the classification options as we went, allowing multiple selections in order to avoid forcing artificial distinctions between stages. The rules are intended to ensure that the lifecycle framework functions as a reproducible coding system, rather than a purely conceptual organizational scheme. The intention is that researchers applying the same selection criteria to the same corpus should arrive at broadly consistent classifications.

\subsection{Methodological Limitations}

The review process has several methodological limitations. The dynamic and developing field of LLM research ensures that new research will come to publication after the search has closed. The inclusion of technical preprints introduces variability in peer-review status; however, such works often represent the leading edge of rapidly evolving LLM research. The classification of studies spanning multiple lifecycle stages also involves interpretive judgment regarding the most appropriate lifecycle categorization. The rules for deciding this process help to clarify the process but do not eliminate subjectivity.

Although the review followed a structured screening and eligibility procedure, the relatively focused corpus selection may limit coverage of newly emerging or highly specialized hallucination techniques, particularly given the rapidly evolving nature of LLM research. The study selection process emphasized methodologically relevant and technically influential works corresponding to major hallucination-related techniques and lifecycle stages. Consequently, additional relevant studies may emerge after completion of the review process.

\subsection{Rationale for the Lifecycle Taxonomy}

The lifecycle taxonomy proposed in this survey organizes hallucination phenomena according to the major operational stages of LLM development and deployment: data, training, and inference. Rather than being directly adopted from a single prior framework, this categorization was developed through an analytical synthesis of recurring patterns identified across the hallucination literature surveyed in this work.

The taxonomy is motivated by the observation that hallucinations emerge through different mechanisms depending on the stage of the LLM pipeline in which instability originates. Data-related factors primarily affect the factual reliability, representational coverage, and informational consistency of the knowledge available during model pretraining. Training-related factors influence how models internalize, optimize, align, and generalize patterns during parameter learning and fine-tuning. Inference-related factors emerge during runtime generation, where decoding strategies, contextual ambiguity, prompt sensitivity, and probabilistic token selection influence output behavior.

The proposed taxonomy is intended as an analytical abstraction rather than a set of strictly isolated mechanisms. In practice, hallucinations may propagate across multiple lifecycle stages. For example, biased or incomplete training data may interact with unstable decoding strategies during inference, producing compounded hallucination effects. Consequently, the framework permits conceptual overlap between categories while preserving stage-specific analysis of hallucination causes, detection mechanisms, mitigation techniques, and prevention strategies.

Unlike prior surveys that primarily organize hallucination research around isolated mitigation or evaluation approaches, the lifecycle framework models hallucination as a propagating phenomenon across interconnected stages of the LLM pipeline. This perspective enables a more systematic understanding of how hallucinations originate, evolve, and can be diagnosed or controlled throughout the lifecycle of LLM development and deployment.

\section{Causes of Hallucination} \label{sec:cause}

Hallucination in LLMs is also a commonly documented phenomenon with multiple sources, which is caused by training data, model design, and inference process \cite{Ji2023,Huang2023, Rawte2023}. It is important to learn about these sources so as to design successful mitigation strategies. An example of this is how GPT-3, when provided with incomplete clinical prompts, has been demonstrated to produce fabricated yet fluent medical prescriptions, demonstrating how hallucinations can be created due to gaps in data, or due to mismatched training, or due to other reasons inference heuristics \cite{Singhal2022}. To address RQ2, this section analyzes the causes of hallucinations across the LLM lifecycle, examining how data, training, and inference conditions contribute to the emergence of hallucinated outputs.

\subsection{Data-Related Causes} \label{subsec:data_related_cause}

Generally, the quality of the training data largely affects the accuracy and factual reliability of the LLM outputs. The weaknesses in the data are usually reflected in the form of hallucinations because the LLMs internalize statistical correlations rather than grounded facts. Some of the important data-related factors which propel hallucination in LLMs are discussed below.

\begin{itemize}
    \item \textbf{Biased Training Data:} Hallucinations can be common when the social or topical bias in the data is incorporated within the LLM \cite{Onoe2022,Sharma2023}. As an example, the bias of LLMs in favor of the Western world is so strong that it occurs even with requests in non-western languages, including the use of Arabic cues to denote western foods (\emph{ravioli}), drinks (\emph{whiskey}), or western female names (\emph{Roseanne}) instead of culturally acceptable ones \cite{Naous2023}. Similarly, comprehensive analyses show that LLM outputs tend to reflect the cultural values of English-speaking, Protestant-European societies, favoring individualistic, Western norms, while marginalizing perspectives from non-Western traditions \cite{Fenech-Borg2025}. This skew reflects the dominant cultural lens of the training data. In addition, \citet{Bender2021} note that large scale internet data encode systemic social and political biases, which models reproduce in generated text. Similarly, \citet{Brown2020} show that GPT-3, when prompted with vague queries, generates racially or gender biased hallucinations (e.g., associating professions with gender), not because such facts are true but because biased data dominate the training distribution.
    
    \item \textbf{Inadequate or Low-Quality Data:} When LLMs are trained on low-quality or unattributed information, they tend to produce contextless or attributionless outputs\cite{Lee2022}. For example, models trained on scraped web text without source attribution may generate factual statements but fail to cite any origin, or worse, fabricate plausible looking references. This phenomenon may be interpreted through parallels with source misattribution or source amnesia in human cognition, where individuals remember information but fail to accurately recall its origin. Other early LLMs, including GPT-2, used uncurated Common Crawl data (which is full of noise, inconsistencies, and spam) to a large degree, further enhancing hallucinations. \citet{Kandpal2022} experiment on factual inconsistency of summarization models and come to the conclusion that corpora like XSum, an abstractive summarization dataset, include misaligning or excessive information abstractive references, which lead to hallucinated summaries. In the same fashion, \citet{Maynez2020} demonstrate that the risks are portrayed by the fact that the use of negligible-quality annotated sets makes abstractive summarization models generate false assertions of downstream generation noisiness at dataset level.
    
    \item \textbf{Ambiguous or Conflicting Data:} Ambiguities of training corpora, e.g. conflicting statements, or of context vague information, may lead to fabricated explanations in LLMs. For example, dialogue models mixed quality conversations trained on tend to yield self contradictory or unsupported statements as is demonstrated by \cite{Shuster2021}. In dialogue tasks based on knowledge grounding, in case there is a lack of supporting evidence or its is unclear, models often fantasize to keep the conversation going. \citet{Zheng2023} also demonstrate that LLMs regresses to poor contextual information in uncertain environments, making false predictions with substantial certainty.
    
\end{itemize}

\subsection{Training-Related Causes} \label{subsec:training_related_cause}

The inefficiencies in the learning process, both in the ability of the model to generalize and to the congruity of its optimization goals to factual correctness, frequently causes hallucinations in LLMs. There are a number of determinants to this phenomenon: 

\begin{itemize} 

    \item \textbf{Overfitting and Memorization:} When a model memorizes certain training examples rather than educating generalizable trends, this is referred to as overfitting \cite{Kang2024,Gekhman2024,Chang2022}. This may take the form of verbatim retraction of factual or non-factual information observed in the course of training in LLMs. This behavior shares certain conceptual similarities with confabulation in human cognition, where coherent but inaccurate details may emerge to fill informational gaps. Indicatively, \citet{Lee2022} demonstrate that GPT-2 is able to replicate long transcribes training text word-to-word, especially with domain specific prompts, and this can cause extremely assertive yet false results. This is particularly worrisome in such sensitive fields like medicine or law, where the memorized information of hallucinations can mislead the users and cause massive damages. 

    \item \textbf{Insufficient or Misaligned Fine-Tuning:} Inappropriate fine-tuning can in itself cause, or worsen, hallucinations. In cases of the fine-tuning of LLMs on small or limited datasets without sufficient alignment to facts, the patterns present in the fine-tuning data may be overemphasized, leading to confidently incorrect or non-consistent results \cite{Singhal2023,Yang2024}. This behavior may exhibit parallels with recency-related cognitive biases in humans, where recently emphasized information disproportionately influences subsequent responses. As an example, \citet{Huang2023} note that LLaMA fine-tuned on small legal corpora occasionally produce hallucinatory legal sources or misuse rules due to the inadequacy of the fine-tuning data opposed to selective or partial. Therefore, a poor or improper fine-tuning can be a source of hallucination instead of a treatment \cite{Zhang2024c}.

    \item \textbf{Architectural Biases:} Some architectural designs and optimization goals may inadvertently concentrate on facts instead of fluency, which makes the models more susceptible to hallucinations \cite{Brown2020}. For instance, decoder only style models e.g. GPT models are optimized to predict the next token conditioned on the past outputs, which may produce very fluent grammatically unsound continuations. Methods that focus on enhancing the likelihood with any grounding, can strengthen the hallucinations. Though this effect is demonstrated by \citet{Du2024} in the text generation tasks only, the principle is applicable in general to LLMs: the well articulated output may include internally inconsistent or artificial information in case the source context of alignment mechanisms are inadequate.

\end{itemize}

\subsection{Inference-Related Causes} 
\label{subsec:inference_related_cause}

Even where the underlying model architecture is robust and is trained on high-quality data, inference can give rise to hallucinations. These are the problems which are usually predetermined by the interpretation of inputs and the creation of outputs. These causes can be generally labeled as, Explanation-Related hallucinations and Classification-Related hallucinations.

\subsubsection{Explanation-Based}

These reasons are connected with the way the model acts on prompts and builds on generative outputs.

\begin{itemize}
    \item \textbf{Input Conflicts:} Inconsistent or ambiguous prompts generally give inconsistent or unstable outputs. For example, a question such as \emph{"Who will be the prime minister of the India in 2027, assuming that the next election will take place in 2026"} has inconsistent time indications which can disorient the model. LLM can also be based on surface information, generating responses which vary on repeated queries, which may resemble forms of contextual suggestibility observed in human decision-making under uncertainty. \citet{Bartsch2023} demonstrate that words that are not clearly articulated may lead to dramatic inconsistencies, and LLMs may occasionally do so contradicting themselves. Also, \citet{Wang2020} and \citet{Liu2023a} emphasize the fact that under-specified questions enhance the risk of hallucinated or false information.
    
    \item \textbf{Contextual Misalignment:} When contextual information is not correctly integrated or retained it may result in hallucinations. The model can produce contextually irrelevant or actually false results when the previous inputs are ignored or wrongly used \cite{Perez2022}. This behavior may share certain interpretive similarities with source misattribution phenomena in human cognition, where recalled information becomes detached from its original context. According to \citet{Chen2023}, deficient use of context has resulted into in hallucinated translations, whereas \citet{Kryciski2019} demonstrate that summarization models generate false statements in case of improper alignment of context.  

    \item \textbf{Extrapolation Beyond Training Data:} Most of the time, LLMs have a high confidence while giving inaccurate responses to inputs that do not belong to the distribution that they trained on. An example is that given the query of the \emph{population of the city of Atlantis}, a model can give a particular number even though the city is not real. This behavior may be heuristically compared to confabulation-like phenomena, where coherent but unsupported responses emerge in the presence of informational uncertainty. \citet{Miao2021} discovered that the prediction of sureness and falsehood in LLMs produces plausible sounding and high internal confidence, indicating the lack of connection between predicted certainty and actual truth.    
\end{itemize}

\subsubsection{Classification-Based} 

These factors are associated with the way the model picks or prioritizes outputs during inference especially in the case of constrained or heuristically-driven decoding strategies.

\begin{itemize}
    \item \textbf{Sampling Optimization and Decoding Errors:} Hallucinations can be increased by inefficient decoding or sampling methods including greedy search, beam search, low-temperature sampling among others \cite{Stahlberg2019, Holtzman2020}. These strategies are high probability based:

    \[
    \hat{t}_i = \arg\max_{t_i} p(t_i \mid t_{<i})
    \]
    in which $\hat{t}_i$ is the predicted token at their position $i$, $p(t_i \mid t_{<i})$ is the conditional probability of token $t_i$ at the preceding sequence $t_{<i}$ and $\arg\max$ is used to choose the token with the highest probability. This is as local as possible, but may spread factual errors throughout the sequence. For example, when generating a biography, the model may select a high-probability but incorrect occupation (e.g., predicting “doctor” instead of “lawyer” for a historical figure) and then consistently build on this error, producing an entirely fabricated narrative \cite{Maynez2020, Min2023}.  

    In \textbf{beam search}, multiple candidate sequences ($B$ beams) are maintained, scoring each sequence $s_j$ by cumulative log-probability:
    \[
    S(s_j) = \sum_{i=1}^{|s_j|} \log p(t_i \mid t_{<i})
    \]
    where $S(s_j)$ is the score of sequence $s_j$, $|s_j|$ is its length, and the summation accumulates the log-probability of each token given its preceding context. Although beam search can reduce some instability, it may still entrench hallucinated outputs if the high-probability sequences happen to be factually wrong \cite{Stahlberg2019, Min2023}.  

    In \textbf{low temperature sampling}, setting ($T < 1$) sharpens the distribution:
    \[
    p_T(t_i \mid t_{<i}) = \frac{p(t_i \mid t_{<i})^{1/T}}{\sum_{t'} p(t' \mid t_{<i})^{1/T}}
    \]
    where $p_T(t_i \mid t_{<i})$ is the adjusted probability under temperature $T$, and the denominator normalizes over all possible tokens $t'$. A lower temperature ($T < 1$) concentrates probability mass on the most likely tokens, reducing diversity but increasing the risk of confidently producing incorrect facts, thereby exacerbating hallucinations \cite{Holtzman2020}. 
    
    Although these decoding strategies optimize token selection based on probabilistic likelihood, they do not explicitly guarantee factual correctness. In greedy decoding and beam search, early token selection errors may propagate through subsequent predictions because each generated token influences future probability distributions. Similarly, low-temperature sampling concentrates probability mass around dominant token candidates, reinforcing associations learned during training. When such associations are biased, weakly grounded, or factually incorrect, the decoding process can amplify them across generation steps, producing fluent yet fabricated outputs that appear internally consistent despite lacking factual validity.
\end{itemize}

\begin{table*}[t] 
\centering
\footnotesize
\caption{Root--Trigger--Manifestation synthesis of hallucination causes across the LLM lifecycle.
This table abstracts individual causes into a unified causal pathway}
\label{tab:rtm_causes}
\resizebox{\textwidth}{!}{
\begin{tabular}{p{1.6cm} p{3.4cm} p{3.8cm} p{4cm} p{1.5cm}}
\toprule
\textbf{Lifecycle Stage} &
\textbf{Root Cause} &
\textbf{Trigger Condition} &
\textbf{Hallucination Manifestation} &
\textbf{Paper} \\
\midrule

\multirow{3}{*}{} &
Biased Training Data &
Ambiguous or culturally sensitive prompts &
Culturally biased or socially skewed generations &
\cite{Bender2021} \\

\addlinespace[5pt]

\textbf{Data} &
Inadequate or Low-Quality Data &
Lack of evidence or source grounding &
Fabricated facts, missing attribution, false citations &
\cite{Maynez2020} \\

\addlinespace[5pt]

 &
Ambiguous or Conflicting Data &
Uncertain context or multi-source conflict &
Self-contradictory or unsupported outputs &
\cite{Shuster2021} \\

\addlinespace[2pt]
\midrule

\multirow{3}{*}{} &
Overfitting and Memorization &
Domain-specific or memorized prompts &
Verbatim recall or confident false continuation &
\cite{Kang2024} \\

\addlinespace[5pt]

\textbf{Training} &
Insufficient or Misaligned Fine-Tuning &
Narrow supervision or biased alignment data &
Overconfident but incorrect responses &
\cite{Yang2024} \\

\addlinespace[5pt]

 &
Architectural Biases &
Long-form generation without grounding &
Fluent yet factually incorrect text &
\cite{Du2024} \\

\addlinespace[2pt]
\midrule

\multirow{4}{*}{} &
Input Conflicts &
Underspecified or contradictory prompts &
Inconsistent answers across prompts &
\cite{Liu2023b} \\

\addlinespace[5pt]

\textbf{Inference} &
Contextual Misalignment &
Long context or poor context usage &
Context-irrelevant or fabricated responses &
\cite{Perez2022} \\

\addlinespace[5pt]

 &
Extrapolation Beyond Training Data &
Out-of-distribution queries &
Highly confident fabricated knowledge &
\cite{Miao2021} \\

\addlinespace[5pt]

 &
Sampling Optimization and Decoding Errors &
Greedy / beam / low-temperature sampling &
Error amplification and hallucination snowballing &
\cite{Holtzman2020} \\

\bottomrule
\end{tabular}}
\end{table*}

\noindent The analysis of hallucination causes across the LLM lifecycle reveals that hallucinations rarely arise from a single isolated mechanism but instead emerge from interacting factors spanning data quality, training dynamics, and inference behavior. The synthesis presented in \autoref{tab:rtm_causes} demonstrates how root causes such as biased or incomplete training data, misaligned training objectives, and unstable inference conditions propagate through specific trigger conditions and ultimately manifest as observable hallucination behaviors. Data-related issues primarily introduce factual gaps and biases that models internalize during pretraining, while training-related factors such as overfitting and architectural biases amplify these weaknesses by encouraging fluent but unsupported continuations. At inference-time, decoding strategies, context limitations, and input ambiguities further exacerbate these latent vulnerabilities. This lifecycle-based synthesis therefore highlights that hallucinations should be viewed as a systemic phenomenon rather than a localized failure, implying that effective solutions must address multiple stages of the model pipeline simultaneously rather than relying on single-stage interventions.

\section{Detection of Hallucination} \label{sec:Detection}

The preceding section outlined multiple causes of hallucinations in LLMs, spanning issues in data, training, and inference-time constraints. However, mitigation requires first identifying and understanding the sources of hallucination. To address RQ3, this section examines methods used to identify hallucinations, focusing on observable indicators and diagnostic techniques across lifecycle stages.

\subsection{Data-Related}

The causes of hallucinations are usually the lack of the training data such as contamination, biased distributions, or otherwise.
lost and inaccurate information. A number of methods used to diagnose these issues include:

\begin{itemize}
  \item \textbf{Data Deduplication and Quality Assessment:} The goal of this is to eliminate training corpora redundancy and noise, minimizing hallucinations on over-represented examples or undesirable examples. Methods include measuring the token Redundancy Rate, Perplexity, and Data Diversity Scores in order to detect the faulty instances. Such preprocessing is commonly used in pre-training or fine-tuning to increase the integrity of data, particularly in large-scale or multilingual models. Indicatively, an example by \citet{Pfeiffer2023} showed that the removal of repetitive sentences and removal of noisy translations of training corpora resulted in a reduced number of hallucinations in machine translation systems, demonstrating that more factual data is cleaner and more balanced makes factual consistency improve directly. Typical preprocessing techniques include similarity-based deduplication using locality-sensitive hashing (LSH), clustering-based outlier detection, statistical anomaly analysis, and heuristic filtering methods for identifying duplicated, noisy, or inconsistent training samples that may reinforce hallucination-prone associations during model training.

 \item \textbf{Truthfulness Benchmarks:} The TruthfulQA \cite{Lin2021} also measures the ability of models to resist the production of aligned outputs with wrong ideas or imaginary opinions. Such metrics as the TruthfulQA Score provide insight into the percentage of responses that are not only informative but correct. These standards willingly bring in misleading or false cues to reveal the weaknesses of a model in the ability to distinguish truth and misinformation. They come in handy especially in opendomain. QA or dialogue systems in which world knowledge is in the center stage. Moderate levels of confidence are below. But since domain-specific hallucinations might not be easily detected as a consequence of the general nature of the benchmarks \cite{Li2023i}.

 \item \textbf{Grounded Fact-Checking:} It compares outputs of the models with the reliable databases like Wikidata or Wikipedia. Metrics that are generally used in this technique are Precision@K \cite{Manning2008}, FactScore \cite{Min2023} and Knowledge Grounded Factuality Score (KG-Fact) \cite{Honovich2021}. These measure the overlap in generated outputs and retrieved facts. This is the most effective way in applications of high stakes and where factual reliability is of paramount importance. This technique is highly effective given that the external knowledge bases are updated on a regular basis. As an example, \citet{Cao2021} and \citet{Li2024i} recorded significant decreases in hallucinations in the combination of adaptive retrieval and fact-grounding. 
 
 \item \textbf{Contextual Consistency Checks:} These approaches evaluate whether generated outputs remain consistent with the conditioning information or source data provided to the model. Mismatches, such as contradictions with background facts, retrieved context, or predefined persona characteristics, may indicate potential hallucinations. Metrics such as Contextual Entailment Accuracy and Dialogue Coherence Score are commonly used to measure this alignment. Although these methods operate on generated outputs, they are categorized under data-related detection within the proposed lifecycle framework because they primarily assess the faithfulness of the generation to the provided grounding data rather than analyzing inference-time decoding behavior. These approaches are particularly useful in multi-turn dialogue and summarization tasks, although their effectiveness may decrease in loosely structured or highly open-ended conversations \cite{Das2023}.

\end{itemize}

\subsection{Training-Related}

Substantial errors added throughout the training process (e.g. overfitting, exposure bias, non-aligned optimization goals, etc.) tend to influence the predictability of the model results. To identify these problems, one will have to look beyond corporate performance but also at finer indicators of information processes as with the model. The approaches below give orderly means of accomplishing when models start becoming unstable and inconsistent.

\begin{itemize}
    \item \textbf{Perplexity and Entropy-Based Detection: }  
    When the models are either too confident about making erroneous judgments or too high changeable, their outputs are subject to instability. It is necessary to quantify the equilibrium between this distrust and doubt in training. Perplexity is a measure of the predictive accuracy of the model on the following token entropy is the dispersion of possible tokens probability. Researchers are able to monitor such measures identify extremes: when perplexity is very low, it indicates overconfidence, and when it is very high, it means that one is confused. These two indications are warning signs, as they indicate that the model is learning distributions which encourage either false fluency or incoherence \cite{Bang2023}. For example, unusually high token entropy during intermediate prediction stages may indicate instability or uncertainty in the model’s internal representations prior to hallucinated generation. Conversely, extremely low entropy may reflect overconfident reliance on narrow token distributions that reinforce weakly grounded associations. Such signals are categorized as training-related indicators because they reflect representational behaviors learned during model optimization rather than only the final generated output.

    \item \textbf{Model Internal State Analysis:}  
    Sometimes the representation of knowledge internally by a model can be distorted due to training making it make erroneous associations. To identify such distortions, one has to look at the hidden layer activations that shape predictions. The lens that internal state analysis offers is the probing of individual neurons or clustering activation patterns to show deviant dynamics. As an example, over-active neurons at all times can be related to the misaligned concepts, where the model attempts to arrive at false answers with confidence \cite{Azaria2023}. This approach highlights representational faults which would otherwise be unseen at the output level. For instance, interpretability analyses may reveal neuron clusters that repeatedly activate for hallucinated entities or unsupported factual associations, thereby exposing internal representational instabilities learned during training.

    \item \textbf{Instruction-Following and Faithfulness Evaluation:}  
    One of the biggest mistakes occurs when models seem to obediently but covertly bending the truth. The model is made to adhere to factual and timely training constancy of behavior is hence very important. This is directly tackled by faithfulness evaluation testing the models remain stable with proven information even when coming up with fluent answers. Measures like FaithDial \cite{Dziri2022} can be used to measure outputs against human labeled truths in order to examine their compliance with instructions as well as factual information. In practice, this assists in finding training arrangements in which the model merely follows the instructions and internalizes distortion of truths below the surface \cite{Adlakha2024}. 

    \item \textbf{Attention Weight Analysis:}  
     A model that puts attention on irrelevant or diffuse portions of the input risks basing guesses of poor associations and not contextual meanings. This compromises the reliability of facts, even in cases where syntax remains correct. Attention weight analysis will give an opportunity to see how the attention is distributed among tokens. With the help of entropy or alignment scores, researchers are able to recognize when training promotes unstable focus patterns \cite{Wu2023}. The signs of semantic drift are evident at one stage of lost attention when the fluency hides a weakening connection to the input. 

    \item \textbf{Layer-Wise Activation Testing}
    Training problems tend to occur biasedly among network layers, having certain layers that increase the instability rather than others. Identifying these layers is important to comprehend how the wrong logic accumulates. Layer-wise activation testing addresses this by measuring \emph{variance across neuron activations:}
    \[
    \sigma_l^2 = \frac{1}{n}\sum_{i=1}^{n}\left(a^{(l)}_i - \mu_l\right)^2, 
    \quad \text{where } \mu_l = \frac{1}{n}\sum_{i=1}^{n} a^{(l)}_i
    \]
    Here, $n$ is the total number of neurons, $\sigma_l^2$ is the variance of activations in layer $l$, $\mu_l$ is the mean activation across neurons in layer $l$, and $a^{(l)}_i$ is the activation of the $i$th neuron in that layer. High values of variance signify big changes in the variables affirmative or unstable layer activity.

    In order to further diagnose abnormal neurons, the \emph{spike detection techniques} are used to compare individual activations with a threshold:
    \[
    a^{(l)}_i > \mu_l + k\sigma_l.
    \]
    where, $k$ is a sensitivity constant. If a neuron’s activation $a^{(l)}_i$ is more than this threshold, it is considered to be disproportionately influential relative to its peers. With this technique, irregularities are highlighted to pinpoint certain layers or neurons on which the training is to be done instabilities build up, and thus the manner in which the unreliability of reasoning or hallucinations can arise and spread is tracked through the network \cite{Sun2022}.  

\end{itemize}

\subsection{Inference-Related}

Hallucinations may also occur during the inference-time despite the presence of strong training. Outputs may not be factual due to the choice of decoding, limited window of context or due to the non-existence of external validation. To deal with this, diagnosis techniques at inference are usually categorized under explanation based and classification based techniques. The former seek to reveal the mechanism behind the existence of hallucinations, whereas the latter seek to dissociate trustworthy and unreliable outputs.

\subsubsection{Explanation-Based Techniques}

The following approaches attempt to explain model behavior through the uncovering of reasoning instabilities leading to hallucinations.

 \begin{itemize}
 
     \item \textbf{Prompt Sensitivity Analysis:} As with slight variations in phrasing, an answer can be entirely different, it is significant to test the stability of a model under immediate variation. This analysis gives this measure by experimenting the behavior of outputs to reword prompts which have semantically equivalent rephrases. The methodology involves the use of measures of the \emph{PromptSensiScore (PSS)} which calculates performance variance among prompt variants on the instance level \cite{Zhuo2024}. Practically it shows fragility in such activities as summarization or open ended QA, hallucinations tend to be aroused by vague prompts. Sensitivity to formatting or sensitivity to format Case studies indicate that sensitivity to format or sensitivity to format exists word choice is closely related to factual drift \cite{Sclar2023,Li2023b}.
     
     \item \textbf{Self Consistency Evaluation:} A model is said to be unstable when it requires different answers to the same question which commonly appears as hallucination. Self-Consistency Evaluation comes up with several of these to capture this output by the use of stochastic decoding algorithms e.g., temperature scaling, top-k, or nucleus sampling \cite{Holtzman2020}. The comparison of these outputs is followed and a Self-Consistency Score calculated as the percentage of logically consistent responses. When the divergence is high, it is a sign of weakness, and when the majority of votes is aggregated it is possible to obtain more credible results \cite{Wang2022}. It is specifically efficient in the exercises that require the ability to think in a systematic way, e.g. math word problems or abstractive summarization, in which unstable lines of reasoning are in general concealed \cite{Zhang2023i}.
     
     \item \textbf{Mechanistic Interpretability Analysis:} Circuits of inference are also prone to errors in inference caused by internal circuits, in which activations, or conversely deactivations, lead to errors in inferences attention is misfire, and these failures are not measurable based on outputs. Mechanistic interpretability addresses it using the internals of the model, including neuron activations, attention heads and layers pathways. The methods of activation patching, attribution of attention, and ablation of neurons assist in the detection of causal linkages between inputs and false outputs. The exposure of such methods is demonstrated in the previous work \cite{Lad2024,Garca-Carrasco2024} clandestine modes of failure and identify the internal pathways to hallucination. This method is very dependable in sensitive areas such as healthcare or law in that it does not only warn about errors but clarifies their occurrence by disclosing the circuits which produce them.
     
 \end{itemize}

\subsubsection{Classification-Based Techniques}

These approaches are aimed at the separation of facts and hallucinated productions with the use of decision rules, external controls or statistical proxies.

 \begin{itemize}
 
     \item \textbf{Cross-Reference Testing:} Hallucinations normally exist since outputs are not compared with known information that is trusted sources. The mitigation of this risk by Cross-Reference Testing is through comparison of model output with external resources (e.g., Wikipedia, domain specific knowledge bases) or verifying consistency with other models. The overlap and consistency are measured by metrics such as \emph{Fact Match Rate (FMR)} and \emph{Cross-Model Agreement Score (CMAS)} \cite{Li2024}. This direct verification is also very useful whereby there is trusted reference data, which provides good guarantee of facts correctness.
     
     \item \textbf{Confidence Score Estimation:} The hallucinations are often found when the model is unsure about something, so to keep track of them, one has to monitor them. The confidence levels can serve as a proxy measure of its confidence levels. Statistical signals used in confidence estimation include as softmax log probability, token level entropy, or uncertainty score calibration. These values are typically learned by summing up the probability of the predicted tokens, i.e. the maximum softmax score given one token, it computes the entropy of the distribution to estimate uncertainty, or the use of a post-hoc calibration techniques including temperature scaling and isotonic regression to optimize the probability estimates. Lower confidence values tend to have incorrect or out of distribution responses. They are not adequate on their part within larger systems of classification, own, these measures are useful constituents, particularly when combined with formulated inference systems that will correct reliability thresholds \cite{Manakul2023}.
     
 \end{itemize}

\begin{figure*}[t]
\centering
\includegraphics[width=0.8\textwidth]{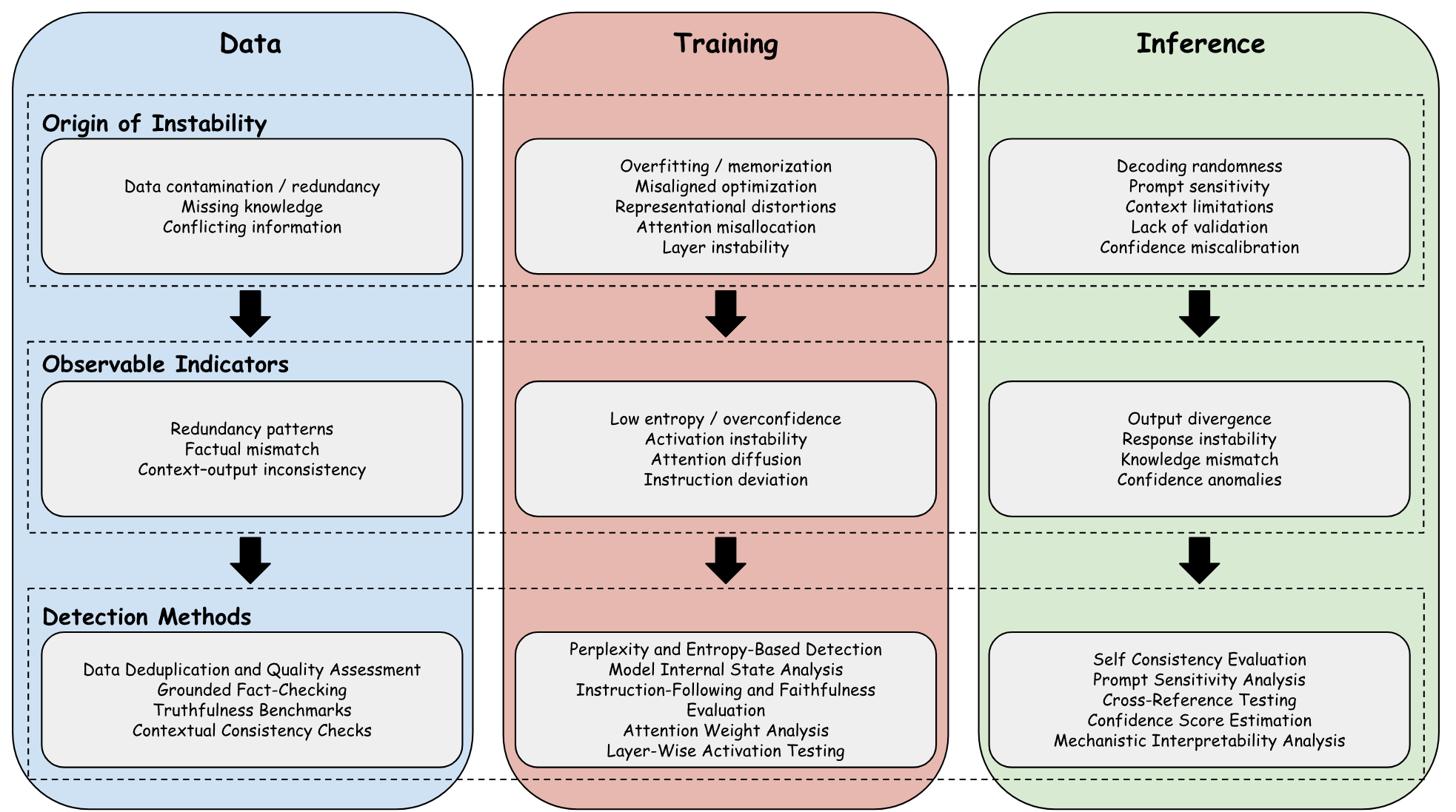}
\caption{Lifecycle-based derivation of hallucination detection methods in LLMs. Each lifecycle stage introduces specific sources of instability that manifest as observable indicators motivating corresponding detection techniques.}
\label{fig:detection}
\end{figure*}

\noindent The lifecycle-based synthesis of detection techniques highlights that hallucination detection methods are fundamentally tied to the observable signals produced at different stages of the LLM pipeline. As illustrated in \autoref{fig:detection}, sources of instability arising from data, training, and inference stages manifest as distinct diagnostic indicators such as redundancy patterns, activation instability, response divergence, or confidence anomalies. Detection techniques therefore operate by translating these signals into measurable diagnostic procedures. Data-level methods typically rely on external verification or benchmark-based evaluation, whereas training-level methods analyze internal model dynamics such as entropy, neuron activations, and attention distributions. Inference-level techniques primarily focus on behavioral consistency and external validation mechanisms. This structured mapping reveals that most existing approaches emphasize output-level verification rather than internal causal diagnosis, suggesting an important research gap in developing methods capable of tracing hallucinations to their internal origins within model representations.

\section{Mitigation of Hallucination} \label{sec:mitigation}

In addition to determining the causes of hallucinations, it is also relevant to discuss the approaches that may help to reduce them so that LLMs will become trustworthy, transparent, and responsible. To address RQ4, this section examines mitigation strategies that aim to reduce or correct hallucinations after they occur, analyzing techniques applied at the data, training, and inference stages of the LLM lifecycle.

In this survey, mitigation refers to interventions designed to reduce, correct, or contain hallucinations after the core model pretraining stage has already occurred. These approaches typically operate during post-training alignment, retrieval augmentation, decoding control, verification, or response correction. Prevention, by contrast, refers to proactive strategies intended to reduce the likelihood of hallucination emergence before or during foundational model development. Preventive approaches primarily focus on improving dataset quality, strengthening training robustness, embedding factual grounding, and introducing structural constraints that reduce hallucination risks earlier in the lifecycle pipeline. Although certain techniques such as reinforcement learning through human feedback (RLHF) may exhibit characteristics of both mitigation and prevention, this survey categorizes them according to their primary operational role within the lifecycle framework.

\subsection{Data-Related}

Mitigation methods based on data address hallucinations caused by biased, incomplete or conflicting data. Such problems as it has been mentioned in \autoref{subsec:data_related_cause} can easily cause models to make-up through the lack of factual foundation. These techniques reinforce the factual correspondence by relying on external sources of knowledge or enhancing the quality of data that has been gathered after the training. We will relate each of the mitigation measures to the root causes, the reasons why it is significant, what it does, and how it achieves its impact.

\begin{itemize}

    \item \textbf{Retrieval-Augmented Generation (RAG)}: The skewed training corpora (C.D.1) and the presence of conflicting information (C.D.3) cause the models to produce unsupported statements because they excessively depend on piecemeal parametric memory. RAG responds to it by making the generative process richer by retrieving it in proven exterior repositories, including curated databases or structured knowledge graphs \cite{Bchard2024, Lewis2021}. This method is essential since even internal representations cannot be sure of factual grounding especially in dynamic fields where knowledge changes at a faster rate. RAG matches outputs with verifiable passages, which minimizes fabrications and contradictions by retrieving and conditioning responses on the current, verifiable passages. Empirical research also confirms this fact through empirical studies that were carried out by \citet{Yin2024} and \citet{Ram2023}, which attest to the fact that real-time retrieval enhances factual correctness and reduces inconsistencies in generation.

    \item \textbf{Post-Training Data Augmentation} Poor or low quality training data (C.D.2) can usually restrict the capacity of a model to internalize factual accuracy, resulting in the repetitive hallucinations across tasks. Post-training augmentation enhances internal knowledge of the model by fine-tuning it on fact checked and high quality data sets. These are usually corrected outputs of previous generations and adversarial examples in which hallucination was detected \cite{Ding2024, Kumar2025}. By being exposed to such controlled and varied examples, such a model slowly learns to prevent repeating past mistakes. The process of iterative correction makes internal representations stronger, as a result, the model can make more reliable and context-consistent answers in further interactions, which makes them more robust and predictive of human behaviors of future interactions~\cite{Chai2025}.

\end{itemize}

\subsection{Training-Related} 

Mitigation strategies based on training can fine-tune the behavior of the model during or just after the fine-tuning stage by changing internal representations, improving response evaluation, and reducing the number of factual inconsistencies. 

\begin{itemize} 

    \item \textbf{Model Editing:} A risk to making the same error over and over again when models overfit to the wrong facts or memorize an incorrect association (C.T.1) is that it is easy to repeat that error later on. Model editing gives a specific method of correcting such problems without complete retraining. The approach can help to localize corrections, which is cost-efficient and leaves the AP unaffected as it is optimized to correct any erroneous output and avoids deterioration of the performance on the tasks that are not related to the specific erroneous output, as the internal parameters are directly updated \cite{Huang2023, Mitchell2022, Chen2023}. This is mainly useful in dynamic fields of knowledge where rapid revisions are necessary. 

    \item \textbf{Attention Reweighting:} Attention mechanisms are usually prone to architectural biases(C.T.3), resulting in misleading or irrelevant tokens which increase the risks on hallucination. This is alleviated by attention reweighting by shifting weights to semantically faithful tokens and thus restructuring the information prioritization of models. Entities aware modulation and deceptive attention detection are techniques shown to reduce distortions in the interpretation of facts and to favor the trustful signals by refining the token-level focus of the model and thus benefiting the model in its purposes \cite{Su2024, Elaraby2023}. 

    \item \textbf{Reinforcement Learning through Human Feedback (RLHF):} When models are undertrained or the objectives are poorly aligned (C.T.2), they are likely to be susceptible to hallucinations, as a result of discrepancies between pretraining tasks and human preferences. RHPF helps in attaining this gap by including human judgments of correctness and coherence in the optimization procedure. Using rewards to encourage outputs that are factually accurate and punishments to discourage misleading replies, RLHF aims to align model outputs with human preferences by encouraging responses that are factually reliable, contextually appropriate, honest, and consistent with grounded information \cite{Christiano2017, openai2022chatgpt, Touvron2023}. This coordination has been shown to be successful in systems deployed on large scale where it has had a significant positive effect in gaining user confidence. 

    \item \textbf{Reinforcement Learning with AI Feedback (RLAIF):} Because RLHF requires human-labeled data, this method is expensive in terms of resources, and models not adequately fine-tuned (C.T.2) are susceptible to hallucinations. This is being solved by RLAIF which replaces human annotators with a teacher model which feeds back preference, thus scaling the alignment process, which can be found in \cite{Lee2024}. Alternatives like direct-RLAIF also minimize the use of reward models, making use of the feedback indicator of existing LLMs. Although it more quickly leads to the factual convergence and consistency, the approach is associated with threats of strengthening the biases or hallucinations transmitted by the teacher model itself.

    \item \textbf{Contrastive Learning for Fact-Consistency:} Due to overfitting (C.T.1), models are not able to differentiate between factual and spurious patterns. This is explicitly addressed by contrastive learning which puts models on the task of factual vs. hallucinated pairs, provoking them to sharpen the boundary between correct and made-up results \cite{Wu2024, Zhang2023contrastive, Chen2023m}. Introducing a greater similarity to correct references and a smaller similarity to incorrect ones causes the model to have a greater discrimination mechanism internally, which diminishes the propensity of the model to make false statements. 

    \end{itemize}

\subsection{Inference-Related} 

The goal of inference-time mitigation methods is to minimize hallucinations at the time of generation, without changing the architecture of the model or its pre-trained parameters. These are real-time genre of intervention strategies and can be grouped into two techniques, which include Explanation-Based and Classification-Based techniques. 

\subsubsection{Explanation-Based Techniques} 
These are techniques that improve the models introspection, analysis, and adaptation of its own answer to more factual accuracy. 

\begin{itemize} 

    \item \textbf{Self-Reflection:} In ambiguous or conflicting situations (C.I.1), unsupported claims tend to be produced by the models. Self-reflection allows this danger to be minimized since it allows the models to review their own responses and creates a new response in case of inconsistencies. The model can detect unreliable content and correct to produce more consistent and fact-congruent output through techniques like automated self-contradiction detection~\cite{Mndler2023}, hallucination self-reflection, etc~\cite{Ji2023m}.

    \item \textbf{Confidence Calibration:} Overconfidence (C.I.2) and Decoding error (C.I.4) in uncertain predictions often lead to hallucinations that look so real. Confidence calibration can solve this by improving the probability distribution of generated responses with the help of temperature scaling and isotonic regression among other methods \cite{Guo2017}. The model detects low confidence estimates as a warning or a recommendation to revise the information to decrease the likelihood of false information being propagated under high confidence by the system, leading to the incorporation of incorrect information that is falsely presented as accurate \cite{Desai2020}.

\end{itemize}

\subsubsection{Classification-Based Techniques}
These are strategies that verify or update generated outputs based on external sources or interactive systems to make sure that they are factually correct. 

\begin{itemize} 

    \item \textbf{intervention:} An intervention with a model can drive it to hallucination due to item conflict (C.I.1) or context conflict (C.I.2), but the interventions in real-time serve as a corrective process. The interventions change the continuous generation to focus on making the outputs and the credible signals similar by use of either reinforcement based adjustments \cite{Li2024m} or adversarial oversight by another model. This response to the hallucination is dynamic correction, which means that a hallucination is detected on the way to the final answer. 
    
    \item \textbf{Fact Checking with External Knowledge Sources:} Extrapolation on data outside of the training set (C.I.3) tends to create fabricated facts that are not grounded in fact. This is minimized by fact checking systems which compare the results with outside, authoritative sources of knowledge during inference. FactLLaMA is a system that combines search engine, database, or language model agent retrieval to confirm claims in real-time \cite{Cheung2023}. Self-Checker is another system that combines all of these to validate claims in real-time \cite{Li2023m}. Relating outputs to external references by anchoring~\cite{Quelle2023} makes these approaches more factual and reduces the spread of falsified or unsubstantiated information. 
\end{itemize}

\begin{table*}[t]
\centering
\footnotesize
\caption{Lifecycle-based quantitative comparison of hallucination mitigation techniques. Causes follow the taxonomy in \autoref{fig:lifecycle}}
\label{tab:mitigation_full}
\resizebox{\textwidth}{!}{
\begin{tabular}{p{1.2cm} p{2.2cm} p{1cm} p{2.2cm} p{3.2cm} p{2.2cm} p{2.8cm} p{1cm}}
\toprule
\textbf{Lifecycle Stage} &
\textbf{Technique} &
\textbf{Cause} &
\textbf{Benchmark / Dataset} &
\textbf{Key Quantitative Outcome} &
\textbf{Computational Cost} &
\textbf{Limitations} &
\textbf{Paper} \\
\midrule

Data &
\textbf{RAG} &
C.D.1, C.D.3 &
NQ, TriviaQA, WebQuestions, CuratedTrec &
NQ EM +10 pts; TriviaQA EM 56.8; FEVER 89.5\% &
High (retrieval + training) &
Retriever dependency; external knowledge bias &
\cite{Lewis2021} \\

\addlinespace[5pt]

 &
\textbf{Post-training Data Augmentation} &
C.D.2 &
BioGEN, LongFact, FollowRAG &
FactScore +8.6; LongFact +5.1; faithfulness +7.2 &
Moderate (offline filtering) &
Single-turn focus; data quality dependency &
\cite{Si2025} \\

\addlinespace[2pt]
\midrule

Training &
\textbf{Model Editing} &
C.T.1 &
VitaminC, QA, FC, ConvSent &
Edit Success 0.986; Drawdown 0.009; stable up to 75 edits &
Moderate (memory modules) &
Scope generalization issues; misuse risk &
\cite{Mitchell2022} \\

\addlinespace[5pt]

 &
\textbf{Attention Reweighting} &
C.T.3 &
WikiBio, NQ, StrategyQA, 2WikiMultiHopQA &
AUC 89.31; StrategyQA acc 0.76; fewer retrieval calls &
Moderate (attention control) &
Depends on token probability access; weak for pretraining hallucinations &
\cite{Su2024} \\

\addlinespace[5pt]

 &
\textbf{RLHF} &
C.T.2 &
Human preference datasets &
Preference win-rate improvement (qualitative) &
High (human feedback) &
Human bias; expensive scaling &
\cite{Christiano2017} \\

\addlinespace[5pt]

 &
\textbf{RLAIF} &
C.T.2 &
Reddit, Helpful-Harmless &
Win-rate 71\%; harmless rate 88\% &
Moderate (AI feedback) &
Teacher model bias; reward model staleness &
\cite{Lee2024} \\

\addlinespace[5pt]

 &
\textbf{Contrastive Learning} &
C.T.1 &
Wizard of Wikipedia &
5--15\% relative gain; $\sim$5$\times$ efficiency &
Moderate (contrastive training) &
Knowledge forgetting; domain generalization &
\cite{Sun2022} \\

\addlinespace[2pt]
\midrule

Inference &
\textbf{Self-reflection} &
C.I.1 &
PubMedQA, MedQuAD, MASH-QA &
MedNLI $\sim$3$\times$ gain; inconsistency 8.69\% $\rightarrow$ 7.38\% &
Moderate (multi-pass inference) &
Domain-specific (medical); not standalone &
\cite{Ji2023m} \\

\addlinespace[5pt]

 &
\textbf{Confidence Calibration} &
C.I.2, C.I.4 &
SNLI, MNLI, QQP, SWAG &
ECE 2.54 $\rightarrow$ 1.14 &
Low (post-hoc scaling) &
Indirect mitigation; task-dependent &
\cite{Desai2020} \\

\addlinespace[5pt]

 &
\textbf{Intervention} &
C.I.1, C.I.2 &
TruthfulQA, NQ, TriviaQA, MMLU &
TruthfulQA +32.6 pts; Vicuna +22.5 pts &
Low (activation control) &
Truthfulness vs helpfulness trade-off &
\cite{Li2023m} \\

\addlinespace[5pt]

 &
\textbf{Fact-checking} &
C.I.3 &
WikiBio &
AUC-PR 93.42; strong factuality correlation &
High (verification calls) &
Computationally heavy; sampling dependent &
\cite{Manakul2023} \\

\addlinespace[2pt]
\bottomrule
\end{tabular}}
\end{table*}

\noindent The comparative analysis of mitigation strategies summarized in \autoref{tab:mitigation_full} indicates that hallucination reduction techniques operate at different levels of the LLM lifecycle and exhibit varying trade-offs between effectiveness, computational cost, and generalizability. Data-level mitigation methods such as retrieval-augmented generation and post-training data augmentation primarily improve factual grounding by incorporating external knowledge or higher-quality training signals. Training-level approaches, including model editing, contrastive learning, and reinforcement learning with human or AI feedback, attempt to modify internal representations to better distinguish factual and spurious patterns. Inference-time methods such as self-reflection, confidence calibration, and external fact-checking act as runtime safeguards that detect or correct unreliable outputs during generation. However, the analysis also reveals that many mitigation techniques rely on additional infrastructure such as retrieval systems, verification pipelines, or feedback models, which introduces new dependencies and computational costs. Consequently, current mitigation strategies function primarily as corrective mechanisms rather than fully eliminating the underlying causes of hallucinations.

\section{Prevention of Hallucination} \label{sec:prevention}

Mitigation techniques deal with correcting what has already happened, whereas prevention techniques deal with the repression of the appearance of hallucinations at the outset. Prevention measures are a proactive solution that lowers the level of post-hoc corrections since they improve the quality of data, reinforce training practices and implement structural model constraints. This is particularly important with health, financial and scientific breakthrough applications, which involve severe consequences of wrong information.

Complementing the mitigation approaches discussed earlier, this section further addresses RQ4 by examining prevention strategies that aim to reduce the likelihood of hallucinations during earlier stages of the LLM lifecycle, including dataset preparation, training design, and inference constraints.

\subsection{Data-Related}

Preventive data measures guarantee that the training corpus is accurate, comprehensive, and well grounded in nature and free of the chances of hallucinating prior to model training.

\begin{itemize}
    \item \textbf{High-Quality Dataset Curation:} This makes sure that the training corpus has correct, properly validated and bias free information and so the model does not learn the wrong facts. Data are subject to severe rigor of data-filtering before training to eliminate misinformation, contradictions and biases. Fact-checking is automated, human annotation, and deduplication are used to verify that the dataset is made up of high quality information that is reliable enough to trust it \cite{Abbas2023}. This helps internalize inaccurate patterns by the model that may subsequently become hallucinations ~\cite{Gunasekar2023}. This is because filtering and curating a top rated textbook style dataset is likely to improve factual consistency, according to \citet{Lee2023}, as a low-quality or speculative source will not be permitted to influence model training. This type of curation enhances factual grounding and results in a significant decrease of the hallucinations during the training and inference.
    
    \item \textbf{Data Diversity \& Coverage Optimization:} Maintaining general diversity and coverage of data is critical in avoiding hallucinations in LLMs. Having a well prepared dataset that encompasses a wide set of topics will reduce the knowledge gaps that could otherwise lead to false assumptions, so although datasets are expected to be diverse in the aspects of topics, linguistic styles, and perspectives, the created content should also be factual and applicable to the task. As demonstrated by \citet{Chung2023}, this is an important factor that should be taken into consideration. Uncontrolled excessive diversity or diversity variation that is not 1:1 with the task goals can create inconsistencies, errors or irrelevant examples, and hence reduce model performance. In the recent past, active learning based approaches have been applied to determine the areas of underrepresented knowledge in order to gather data that can be used to address these knowledge gaps. A framework proposed by \citet{Lin2024} involves the diversity aware active learning that utilizes hallucinations to select a variety of hallucinations that can be annotated and hence the ability of the model to become more robust. With the organized increase in data diversity and more effective coverage, the chances of the LLMs generating hallucinated information due to a small amount of exposure are significantly reduced \cite{Si2025}. These plans enhance resilience because they decrease topic specific hallucinations especially in areas that have lacked sufficient representation.
    
    \item \textbf{Knowledge Grounding:} Knowledge grounding makes sure the generated text is consistent with established bodies of knowledge or other sources of knowledge~\cite{Liu2023b}. Other strategies include the Neural Knowledge Bank of Pretrained Transformers~\cite{Dai2022} where factual knowledge is dynamically injected during generation in that it retrieves and combines the necessary evidence at runtime in addition to just using pre-trained parameters. This is intimately allied to RAG though can include entity linking or knowledge retrieval specialized modules. Moreover, the authors of both articles suggest automated feedback mechanisms that compare the outputs to external resources and amend hallucinations \cite{Peng2023,Yu2023}. This kind of feedback is mainly gathered via automated verification pipelines, yet human in the loop strategies are used in case of low confidence cases selectively only \cite{Cui2024,Gunjal2024}. These grounding strategies enhance the factual accuracy by making sure that brought about responses are verifiable and range of hallucinations in the knowledge intensive areas like science, medicine and law.

\end{itemize}

\subsection{Training-Related}

The preventive methods based on training incorporate limits and verification measures into the parameter learning process to reduce the quantity of hallucination generated content.

\begin{itemize}
    \item \textbf{Fine-Tuning:} Before deployment, fine-tuning on domain-specific and high-quality training data helps models learn accurate and contextually relevant information rather than relying on noisy or potentially unreliable web-crawled data. For example, Llama 2 is fine-tuned on filtered and curated datasets, improving factual accuracy and reducing hallucination tendencies \cite{Touvron2023}. Similarly, reinforcement learning based approaches such as BATGPT \cite{Li2023a} optimize model behavior using reward signals derived from factual correctness and user preferences. By aligning model outputs with reliable domain knowledge and human feedback, fine-tuning reduces the likelihood of generating unsupported or fabricated information, thereby serving as an important preventive mechanism against hallucinations.

    \item \textbf{Adversarial Training:} This procedure increases the robustness against LLMs by introducing adversarial prompts, e.g. counterfactual or misleading prompts, and, in this way, proving the decision boundaries to be more robust~\cite{Zhang2023}. It was based on adversarial robustness techniques in deep learning borrowed by NLP to minimize errors in facts when faced with difficult input~\cite{Madry2017}. Indicatively, to enhance coverage and attack success in text classification, the authors of the study by \citet{Ren2020} suggest creating adversarial text with no input with the help of generative models (e.g., VAE with GAN like training), which enhances the results of text classification. In more recent work, \citet{Yao2023} redefine hallucinations as adversarial examples in their own right, they demonstrate that simple prompt-based hallucination inductions can be reliably obtained with LLMs, and that adversarial, inspired hallucination attacks can be defended by simple defense mechanisms. Adversarial training allows models to be less sensitive to edge cases and adversarial examples and minimizes the prevalence of hallucinations on adversarial examples or noisy examples.
    
    \item \textbf{Neurosymbolic AI:} This is a combination of both neural networks and symbolic reasoning to aid higher fact based accuracy and reduced hallucination during training. These hybrid models directly incorporate structured knowledge, such as ontologies or logical rules, into the learning process to control model updates and constrain the generation of hallucinated patterns in order to create such hybrid models \cite{Sheth2023}. Neurosymbolic systems can instruct more reliable in-house representations by imposing logical constraints and grounding knowledge graphs on the training goals, avoiding any possibility of hallucinating before deployment to the real-world ~\cite{Garcez2023}. This proactive integration is particularly crucial in those applications where control of model behavior is heavily valued, like in question answering and medical reasoning, in which early control is much more useful in enhancing downstream reliability \cite{Colelough2025}. Such a strategy enhances factual endorsement, generating more credible reasoning paths and decreasing domain specific hallucinations of high stakes tasks.
    
    \item \textbf{Knowledge Injection:} Knowledge Injection is the addition of structured, validated and domain specific knowledge into the model during pretraining time and is done to facilitate factual accuracy. This approach includes factual source of knowledge as a direct part of the model whereas many other methods rely on the patterns observed in large, noisy datasets. Other techniques like the Wake-Sleep Algorithm that improve unsupervised learning include cycling between a wake state (inference) and a sleep state (generative learning) to improve internal knowledge representations \cite{Hinton1995}. DreamCoder infuses knowledge by building and optimizing symbolic program like structures, which allow the model to generalize and map factual concepts to interpretable forms of expression \cite{Ellis2020}. Knowledge can be injected by SleepFM and MSSC-BiMamba in a multimodal fashion, that is, physical data (e.g., brain activity, ECG, respiratory signals) are structured and combined with neural representations that restrict the model to provide outputs that are consistent with validated medical signals and, therefore, reducing hallucinations in healthcare tasks \cite{Thapa2024,Zhang2024p}. Knowledge injection results in a greater factual adherence and makes models less dependent on noisy parametric memory and reduces hallucination in specialized fields like science and healthcare.

\end{itemize}

\begin{table*}[!t]
\centering
\footnotesize
\caption{Lifecycle-based comparison of hallucination prevention techniques.
Causes follow the taxonomy in ~\autoref{fig:lifecycle}}
\label{tab:prevention_full}
\resizebox{\textwidth}{!}{
\begin{tabular}{p{1.2cm} p{2.2cm} p{1cm} p{2.2cm} p{3.2cm} p{2.2cm} p{2.8cm} p{1cm}}
\toprule
\textbf{Lifecycle} &
\textbf{Technique} &
\textbf{Cause} &
\textbf{Benchmark / Dataset} &
\textbf{Key Quantitative Outcome} &
\textbf{Computational Cost} &
\textbf{Limitations} &
\textbf{Paper} \\
\midrule

Data &
\textbf{Dataset Curation} &
C.D.1, C.D.2 &
C4, RealNews, LM1B, Wiki-40B &
Memorization $\downarrow$ $\sim$10$\times$ (1.926\% $\rightarrow$ 0.189\%); perplexity improved &
High (data deduplication) &
Heavy preprocessing; may remove useful repetitions &
\cite{Lee2022} \\

\addlinespace[5pt]

 &
\textbf{Data Diversity \& Coverage} &
C.D.1 &
NaturalQuestions, TriviaQA, The Pile &
QA accuracy $\sim$25\% $\rightarrow$ $>$55\% &
Very High (data scaling) &
Scaling alone insufficient; imperfect relevance estimation &
\cite{Kandpal2022} \\

\addlinespace[5pt]

 &
\textbf{Knowledge Grounding} &
C.D.3 &
WikiText-103, RealNews, NQ, TriviaQA &
Perplexity $\downarrow$; factual QA performance improved &
Moderate (retrieval indexing) &
Latency overhead; limited context window &
\cite{Ram2023} \\

\addlinespace[2pt]
\midrule

Training &
\textbf{Fine-Tuning} &
C.T.2 &
TruthfulQA, MMLU, GSM8K &
Truthfulness 50.18 $\rightarrow$ 64.14; toxicity 24.6\% $\rightarrow$ 0.01\% &
Very High (full training) &
Expensive; language bias; benchmark limits &
\cite{Touvron2023} \\

\addlinespace[5pt]

 &
\textbf{Adversarial Training} &
C.T.1 &
MNIST, CIFAR10 &
Robust accuracy 89.3\% under PGD attack &
High (adversarial optimization) &
Optimization difficulty; robustness may not transfer to NLP fully &
\cite{Madry2017} \\

\addlinespace[5pt]

 &
\textbf{Neurosymbolic AI} &
C.T.3 &
Conceptual evaluation + mental health pipeline &
Expert satisfaction 47\% $\rightarrow$ 70\% &
Moderate (symbolic integration) &
Mostly conceptual; limited large-scale benchmarks &
\cite{Sheth2023} \\

\addlinespace[5pt]

 &
\textbf{Knowledge Injection} &
C.T.2, C.T.3 &
Program induction, physical laws tasks &
Task success 3.7\% $\rightarrow$ 79.6\%; search time 235s $\rightarrow$ 40s &
High (knowledge integration) &
Symbolic assumptions; limited real-world noise handling &
\cite{Ellis2020} \\

\addlinespace[2pt]
\midrule

Inference &
\textbf{Prompt Engineering} &
C.I.1 &
Concept-7 dataset &
AUC 0.927; ACC 0.868 &
Low (prompt design) &
Sensitive to prompt style and model family &
\cite{Luo2023} \\

\addlinespace[5pt]

 &
\textbf{Self-Consistency Decoding} &
C.I.2 &
GSM8K, SVAMP, StrategyQA &
Reasoning accuracy +17.9\% (GSM8K) &
Moderate (multi-sampling) &
Slow; generates inconsistent paths &
\cite{Wang2022} \\

\addlinespace[5pt]

 &
\textbf{Controlled Decoding} &
C.I.3, C.I.4 &
CNN-DM, XSUM, MemoTrap, NQ-Swap &
FactKB +14.3\%; EM up to 2.9$\times$ &
Low--Moderate (decoding control) &
Needs context quality; no training adaptation &
\cite{Shi2023} \\

\bottomrule
\end{tabular}}
\end{table*}

\subsection{Inference-Related}

Preventive methods Inference-time preventive methods are designed to generate the outputs in a way that minimizes the occurrence of hallucination-prone outputs prior to their occurrence. These techniques can be broadly classified into two groups namely, explanation-based and classification-based technique.

\subsubsection{Explanation-Based Techniques}

These methods render the generation procedure more explainable and predictable as the model is motivated to rationale across numerous conceivable courses of action or follow clear directions.

 \begin{itemize}
 
    \item \textbf{Prompt Engineering:} Prompt engineering does limit the language model output space to minimise responses associated with hallucinations. Previous research indicates that well spacing out prompts can induce models to factuality by training them to answer only when they are sure or to give supporting evidence~\cite{Luo2023,Rawte2024}. Prompting strategies of metacognition, as proposed by \cite{Wang2024} also enhance the accuracy of responses given by prompting the model to review the reasoning and shun the speculative products. \cite{Barkley2024} that offer systematic evaluation of prompting methods in the reduction of hallucinations. When combined, these strategies make the model outcomes more reliable and reliable as it guides the outputs by responding to verifiable and evidence-based answers that result in fewer speculative generations.
    
    \item \textbf{Self-Consistency Decoding:} It is a very good method of enhancing the reliability of output, as several independent answers to an identical prompt are produced and the most consistent or often occurring response is chosen. This aggregation process filters out inconsistent or outlier generations which are more likely to be affected by hallucination. Self-consistency enhances grounding in fact when it is used together with chain-of-thought prompting where the convergent paths of reasoning are arrived at~\cite{Wang2022}. Recent efforts by~\citet{Wan2024} suggest variants that are reasoning aware and dynamically reason about the quality of rationale, and can also do early stopping. They proposed RASC(Reasoning Aware Self Consistency), which achieved approximately 70\% reduction in sample usage while still maintaining the accuracy. Generally, the outputs of self-consistency decoding are more consistent, verifiable and less prone to hallucinations.

\end{itemize}

\subsubsection{Classification-Based Techniques}  

The methods add auxiliary classifiers or scoring modules into the decoding mechanism to provide structured constraints which force factual correctness. Structured constraints Structured constraints are classification indicators, e.g. factuality score, token plausibility or entailment plausibility, that the model uses to prefer reliable continuations to hallucinated ones. The model is trained to distinguish between good and bad token candidates by decoding it as a constrained classification problem at each step.

\begin{itemize}  
    \item \textbf{Controlled Decoding:} The controlled decoding is a technique that results in restriction of the generation by conditioning it to an external signal or classifier feedbacks. Indicatively, a model can be based on scores of factuality of auxiliary discriminators or verification modules. DoLa (Decoding by Contrasting Layers) make comparisons between candidate-outputs across transformer layers and choose the most consistent response~\cite{Chuang2024}. Nucleus sampling and temperature scaling are other stochastic decoding methods used to control the rates of hallucinations, varying the degree of randomness in the generation~\cite{Holtzman2020,Meister2022}; adjustment of such parameters allows to balance between the degree of factual accuracy and the degree of creativity~\cite{Chang2023}. In other algorithms, probabilistic scoring functions are used, which give greater importance to contextually relevant and plausible tokens, and decrease the chance of generating content that is factually incorrect~\cite{Shi2023,Wan2023}. All of these decoding limitations result in more factual and less fabricated outputs. 
\end{itemize}

\noindent The lifecycle-based comparison of prevention strategies summarized in \autoref{tab:prevention_full} highlights the importance of addressing hallucination risks at earlier stages of the LLM development pipeline. Unlike mitigation techniques, which correct hallucinations after they occur, prevention strategies aim to reduce the probability of hallucination generation by improving dataset integrity, strengthening training procedures, and introducing structural constraints during inference. Data-level approaches such as high-quality dataset curation, diversity-aware data collection, and knowledge grounding focus on eliminating factual gaps and biases before model training begins. Training-level strategies including fine-tuning, adversarial training, neurosymbolic integration, and knowledge injection aim to embed stronger factual reasoning capabilities directly within model representations. Inference-level methods such as prompt engineering and controlled decoding introduce generation constraints that guide models toward more reliable outputs. The comparative analysis in Table 4 also reveals that prevention strategies often require significant computational resources and curated datasets but provide a more sustainable pathway toward reducing hallucination risk by addressing the root conditions that enable hallucinations to emerge.

\section{Suitability Analysis of Hallucination Benchmarks}
\label{sec:benchmark}

As LLMs become increasingly integrated into domains such as healthcare, law, governance, and education, the risk of hallucinations, including factually incorrect, fabricated, or unjustifiably confident outputs, has emerged as a major reliability concern. Evaluating these risks requires benchmark frameworks capable not only of identifying hallucinations, but also of analyzing their causes, mitigation strategies, and control mechanisms across different stages of the LLM lifecycle. To address RQ5, this section examines existing hallucination benchmarks and evaluates their suitability within the proposed lifecycle framework.

\subsection{Benchmark Requirements for Identifying Hallucination Causes}

In order to comprehend the mechanics of hallucinations, the benchmarks should evaluate both factual accuracy and the reasoning processes underlying model outputs. Among the paramount parameters is the \emph{\textbf{availability of ground-truth}} (that is, datasets are provided with reference answers that are authoritative). In the absence of sound ground truth, it will be unclear whether a certain creativity is legitimate or a real hallucination. \emph{\textbf{Input diversity}} is also another significant factor. Formal and informal writing, use of prompts that can be paraphrased, and content rich in domain can demonstrate particular linguistic or contextual circumstances in which hallucinations have higher chances of occurring. \emph{\textbf{Annotations of model responses}}, in which outputs are factual, partially correct, or hallucinated are equally important. Such structured annotations enable more systematic and fine-grained analysis of hallucination behavior. \emph{\textbf{Linguistic and contextual variability}}, including tasks with the ambiguity of the pronouns or cumulative reasoning, also contribute to determining that hallucinations occur when contextual evidence changes in a subtle way. Lastly, benchmarks need to consider \emph{\textbf{reasoning chains or explanations}} and not final answers alone. Corrupt logic processes are usually the sources of premature hallucination even when the given end product is merely superficial.

\subsection{Benchmark Requirements for Evaluating Mitigation Strategies}

In addition to the recognition of hallucinations, benchmarks should be used to determine the effectiveness of various strategies to minimize or eliminate them. This starts with the analysis of the effect of \emph{\textbf{RAG}} wherein the external sources of knowledge are incorporated so that the model is based on facts. Benchmarks, which incorporate \emph{\textbf{error-correction annotations}}, pairs of incorrect and corrected outputs, assist in the determination of the capability of a model to self-revise or to use feedback successfully. The other significant dimension is the strength of outputs to initiate variations promptly with help of \emph{\textbf{prompt engineering and paraphrasing}} tests. Coherence in various wordings is a sign of more factual support. The datasets based on \emph{\textbf{feedback}}, in which models are given rewards based on accuracy and punished based on hallucinations, generate simulated real-world learning and aid in measuring reinforcement-based mitigation. Last but not least, \emph{\textbf{cross-model comparisons}} in the same benchmarking setup allow to directly compare the resilience of what architectures or fine-tuning strategies are more resilient to hallucinations.

\subsection{Benchmark Requirements for Hallucination Control and Monitoring} 

Hallucinations cannot be completely eliminated in real-world deployment settings, so real-world systems must have a system to regulate or contain it. It is thus in the interest of benchmarks to determine how well models estimate and communicate risk by looking at \emph{\textbf{uncertainty measures and confidence scores}}. Uncertainty can be easily estimated and this enables the downstream systems to be aware of possible danger or lack of trust in answers. Likewise, the \emph{\textbf{severity levels of hallucinations}} should be taken into consideration when setting benchmarks and to be able to distinguish between minor anomalies and dangerous factual error to intervene in a prioritized manner. Hallucination control also depends on the model’s ability to estimate and communicate uncertainty, which is based on the assessment of \emph{\textbf{intervention strategies}}, including absence of answering, invoking a retrieval module, or using symbolic rules. \emph{\textbf{Instruction-tuning or policy restrictions}} are complementary in that they can tend to test whether explicit rules, domain constraints or ethical considerations are sufficiently effective in directing the model to avoid responding to hallucination-prone situations. Lastly, there should be \emph{\textbf{human control and adversarial testing}}. Adversarial, ambiguous or deliberately confusing prompts stress testing the model, together with human evaluation of the failures, make sure that the subtle hallucination paths not found by automated measures are detected.

\begin{table*}[!t]
  \centering
  \resizebox{\textwidth}{!}{
  \rotatebox{-90}{
  \begin{minipage}{\textheight}
   \centering
   \caption{Suitable Benchmark Dataset for Hallucination task; $\times$ = Not Suitable, \checkmark = Moderately Suitable, \checkmark\checkmark = Highly Suitable}
   \footnotesize
    \begin{tabularx}{\textwidth}{c|XXXXX|>{\columncolor{gray!30}}X|XXXXX|>{\columncolor{gray!30}}X|XXXXX|>{\columncolor{gray!30}}X}
        \toprule
          & \multicolumn{6}{|c|}{\cellcolor{gray!30} \textbf{Identification of the Causes}} & \multicolumn{6}{|c|}{\textbf{\cellcolor{gray!30}Mitigation of Hallucination}} & \multicolumn{6}{|c|}{\textbf{\cellcolor{gray!30}Enforcing Control over Hallucination}} \\
        \cmidrule{2-19}
         \textbf{Datasets} & \rotatebox{90}{\shortstack{Ground Truth \\ Availability}} & \rotatebox{90}{\shortstack{Diversity of \\ Inputs}} & \rotatebox{90}{\shortstack{Model Response \\ Annotations}} & \rotatebox{90}{\shortstack{Linguistic \& \\ Contextual Variability}} & \rotatebox{90}{\shortstack{Reasoning Chains \\ \& Explanations}} & \rotatebox{90}{Suitability} & \rotatebox{90}{\shortstack{Retrieval-Augmented \\  Data}} & \rotatebox{90}{\shortstack{Error Correction \\ Annotations}} & \rotatebox{90}{\shortstack{Prompt Engineering \\ \& Paraphrasing}} & \rotatebox{90}{\shortstack{Feedback \& \\ Reinforcement Data}} & \rotatebox{90}{\shortstack{Cross-Model \\ Comparisons}} & \rotatebox{90}{Suitability} & \rotatebox{90}{\shortstack{Confidence Scores \& \\ Uncertainty Measures}} & \rotatebox{90}{\shortstack{Hallucination \\ Severity Levels}} & \rotatebox{90}{\shortstack{Intervention \\ Strategies}} & \rotatebox{90}{\shortstack{Instruction-Tuning \& \\ Policy Constraints}} & \rotatebox{90}{\shortstack{Human Oversight \& \\ Adversarial Testing}} & \rotatebox{90}{Suitability}\\
         \midrule
          \cellcolor{gray!30}TruthfulQA & \checkmark &	\checkmark &	\checkmark &	\checkmark &	$\times $ &	\checkmark &	$\times $ &	$\times $ &	\checkmark &	$\times $ &	\checkmark &	$\times $ &	$\times $ &	\checkmark &	$\times $ &	\checkmark &	\checkmark &	\checkmark \\
          \midrule
          \cellcolor{gray!30}TruthfulQA++   & \checkmark &	\checkmark &	\checkmark &	\checkmark &	\checkmark &	\checkmark \checkmark &	$\times $ &	\checkmark &	\checkmark &	\checkmark &	\checkmark &	\checkmark &	\checkmark &	\checkmark &	\checkmark &	\checkmark &	\checkmark &	\checkmark \checkmark  \\
          \midrule
          \cellcolor{gray!30}FEVER   & \checkmark &	\checkmark &	\checkmark &	$\times $ &	$\times $ &	\checkmark &	\checkmark &	\checkmark &	$\times $ &	$\times $ &	$\times $ &	$\times $ &	\checkmark &	$\times $ &	\checkmark &	$\times $ &	$\times $ &	$\times $  \\
          \midrule
          \cellcolor{gray!30}FactCC   & \checkmark &	$\times $ &	\checkmark &	$\times $ &	$\times $ &	$\times $ &	\checkmark &	\checkmark &	$\times $ &	\checkmark &	$\times $ &	\checkmark &	\checkmark &	\checkmark &	\checkmark &	\checkmark &	$\times $ &	\checkmark  \\
          \midrule
          \cellcolor{gray!30}TabFact   & \checkmark &	\checkmark &	\checkmark &	\checkmark &	\checkmark &	\checkmark \checkmark &	\checkmark &	\checkmark &	$\times $ &	$\times $ &	$\times $ &	$\times $ &	\checkmark &	\checkmark &	$\times $ &	$\times $ &	$\times $ &	$\times $ \\
          \midrule
          \cellcolor{gray!30}CounterFact   & \checkmark &	\checkmark &	\checkmark &	\checkmark &	\checkmark &	\checkmark \checkmark &	\checkmark &	\checkmark &	$\times $ &	$\times $ &	\checkmark &	\checkmark &	\checkmark &	\checkmark &	\checkmark &	$\times $ &	\checkmark &	\checkmark \\
          \midrule
          \cellcolor{gray!30}Wikidata  & \checkmark &	$\times $ &	\checkmark &	\checkmark &	$\times $ &	\checkmark &	\checkmark &	\checkmark &	$\times $ &	$\times $ &	$\times $ &	$\times $ &	\checkmark &	$\times $ &	\checkmark &	$\times $ &	$\times $ &	$\times $  \\
          \midrule
          \cellcolor{gray!30}XSum  & \checkmark &	\checkmark &	\checkmark &	\checkmark &	$\times $ &	\checkmark &	$\times $ &	\checkmark &	\checkmark &	$\times $ &	\checkmark &	\checkmark &	$\times $ &	\checkmark &	\checkmark &	$\times $ &	\checkmark &	\checkmark \\
          \midrule
          \cellcolor{gray!30}HaluEval & \checkmark &	\checkmark &	\checkmark &	\checkmark &	\checkmark &	\checkmark \checkmark &	$\times $ &	\checkmark &	\checkmark &	\checkmark &	\checkmark &	\checkmark &	\checkmark &	\checkmark &	\checkmark &	\checkmark &	\checkmark &	\checkmark \checkmark  \\
          \midrule
          \cellcolor{gray!30}HaDeS & \checkmark &	\checkmark &	\checkmark &	\checkmark &	\checkmark &	\checkmark \checkmark &	\checkmark &	\checkmark &	\checkmark &	$\times $ &	\checkmark &	\checkmark &	\checkmark &	\checkmark &	\checkmark &	\checkmark &	\checkmark &	\checkmark \checkmark  \\
          \midrule
          \cellcolor{gray!30}FactCHD & \checkmark &	\checkmark &	\checkmark &	\checkmark &	\checkmark &	\checkmark \checkmark &	\checkmark &	\checkmark &	$\times $ &	\checkmark &	$\times $ &	\checkmark &	\checkmark &	\checkmark &	\checkmark &	$\times $ &	\checkmark &	\checkmark  \\
          \midrule
          \cellcolor{gray!30}RAGTruth & \checkmark &	\checkmark &	\checkmark &	\checkmark &	\checkmark &	\checkmark \checkmark &	\checkmark &	\checkmark &	\checkmark &	\checkmark &	\checkmark &	\checkmark \checkmark &	\checkmark &	\checkmark &	\checkmark &	\checkmark &	\checkmark &	\checkmark \checkmark \\
          \midrule
          \cellcolor{gray!30}FaithDial & \checkmark &	\checkmark &	\checkmark &	\checkmark &	$\times $ &	\checkmark &	\checkmark &	\checkmark &	$\times $ &	\checkmark &	$\times $ &	\checkmark &	\checkmark &	\checkmark &	\checkmark &	\checkmark &	$\times $ &	\checkmark  \\
          \midrule
          \cellcolor{gray!30}CNN/Daily Mail & \checkmark &	\checkmark &	\checkmark &	$\times $ &	$\times $ &	\checkmark &	$\times $ &	$\times $ &	\checkmark &	$\times $ &	\checkmark &	$\times $ &	$\times $ &	\checkmark &	$\times $ &	$\times $ &	$\times $ &	$\times $  \\
          \midrule
          \cellcolor{gray!30}WebGPT  & \checkmark &	\checkmark &	\checkmark &	\checkmark &	\checkmark &	\checkmark \checkmark &	\checkmark &	\checkmark &	\checkmark &	\checkmark &	\checkmark &	\checkmark \checkmark &	\checkmark &	\checkmark &	\checkmark &	\checkmark &	\checkmark &	\checkmark \checkmark \\
          \midrule
          \cellcolor{gray!30}HypoTermQA & \checkmark &	\checkmark &	\checkmark &	\checkmark &	\checkmark &	\checkmark \checkmark &	$\times $ &	\checkmark &	\checkmark &	$\times $ &	\checkmark &	\checkmark &	\checkmark &	$\times $ &	\checkmark &	$\times $ &	\checkmark &	\checkmark  \\
          \midrule
          \cellcolor{gray!30}DELUCIONQA & \checkmark &	\checkmark &	\checkmark &	\checkmark &	\checkmark &	\checkmark \checkmark &	\checkmark &	\checkmark &	\checkmark &	\checkmark &	\checkmark &	\checkmark \checkmark &	\checkmark &	\checkmark &	\checkmark &	\checkmark &	\checkmark &	\checkmark \checkmark  \\
          \midrule
          \cellcolor{gray!30}DefAn & \checkmark &	\checkmark &	\checkmark &	\checkmark &	\checkmark &	\checkmark \checkmark &	$\times $ &	\checkmark &	\checkmark &	$\times $ &	$\times $ &	$\times $ &	\checkmark &	\checkmark &	$\times $ &	$\times $ &	$\times $ &	$\times $  \\
          \midrule
          \cellcolor{gray!30}HalluDial & \checkmark &	\checkmark &	\checkmark &	\checkmark &	$\times $ &	\checkmark &	$\times $ &	\checkmark &	\checkmark &	\checkmark &	$\times $ &	\checkmark &	\checkmark &	\checkmark &	\checkmark &	\checkmark &	$\times $ &	\checkmark \\
          \midrule
          \cellcolor{gray!30}PubMedQA & \checkmark &	\checkmark &	\checkmark &	\checkmark &	\checkmark &	\checkmark \checkmark &	\checkmark &	\checkmark &	$\times $ &	$\times $ &	$\times $ &	$\times $ &	\checkmark &	\checkmark &	\checkmark &	$\times $ &	$\times $ &	\checkmark \\
          \midrule
          \cellcolor{gray!30}BioASQ & \checkmark &	\checkmark &	\checkmark &	\checkmark &	\checkmark &	\checkmark \checkmark &	\checkmark &	\checkmark &	$\times $ &	\checkmark &	$\times $ &	\checkmark &	\checkmark &	\checkmark &	\checkmark &	\checkmark &	$\times $ &	\checkmark \\
          \midrule
          \cellcolor{gray!30}MS MARCO & \checkmark &	\checkmark &	\checkmark &	\checkmark &	$\times $ &	\checkmark &	\checkmark &	$\times $ &	\checkmark &	$\times $ &	\checkmark &	\checkmark &	\checkmark &	$\times $ &	\checkmark &	$\times $ &	\checkmark &	\checkmark \\
          \midrule
          \cellcolor{gray!30}PersonaChat & \checkmark &	\checkmark &	\checkmark &	\checkmark &	$\times $ &	\checkmark &	$\times $ &	$\times $ &	\checkmark &	\checkmark &	$\times $ &	$\times $ &	$\times $ &	\checkmark &	$\times $ &	\checkmark &	$\times $ &	$\times $ \\
          \midrule
          \cellcolor{gray!30}OpenDialKG & \checkmark &	\checkmark &	\checkmark &	\checkmark &	$\times $ &	\checkmark &	\checkmark &	\checkmark &	\checkmark &	\checkmark &	$\times $ &	\checkmark &	\checkmark &	\checkmark &	\checkmark &	\checkmark &	$\times $ &	\checkmark \\
          \midrule
          \cellcolor{gray!30}DialogSum & \checkmark &	\checkmark &	\checkmark &	$\times $ &	$\times $ &	\checkmark &	$\times $ &	\checkmark &	\checkmark &	$\times $ &	\checkmark &	\checkmark &	$\times $ &	\checkmark &	\checkmark &	$\times $ &	\checkmark &	\checkmark  \\
          \midrule
          \cellcolor{gray!30}MultiWOZ & \checkmark &	\checkmark &	\checkmark &	\checkmark &	$\times $ &	\checkmark &	$\times $ &	\checkmark &	\checkmark &	\checkmark &	\checkmark &	\checkmark &	$\times $ &	\checkmark &	\checkmark &	\checkmark &	\checkmark &	\checkmark \\
          \midrule
          \cellcolor{gray!30}HALBench & \checkmark &	\checkmark &	\checkmark &	\checkmark &	\checkmark &	\checkmark \checkmark &	$\times $ &	\checkmark &	\checkmark &	\checkmark &	\checkmark &	\checkmark &	\checkmark &	\checkmark &	\checkmark &	\checkmark &	\checkmark &	\checkmark \checkmark  \\
          \midrule
          \cellcolor{gray!30}CommonGen & \checkmark &	\checkmark &	\checkmark &	\checkmark &	$\times $ &	\checkmark &	$\times $ &	\checkmark &	\checkmark &	$\times $ &	\checkmark &	\checkmark &	$\times $ &	\checkmark &	\checkmark &	$\times $ &	\checkmark &	\checkmark \\
          \midrule
          \cellcolor{gray!30}TREX & \checkmark &	\checkmark &	\checkmark &	\checkmark &	$\times $ &	\checkmark &	\checkmark &	$\times $ &	$\times $ &	$\times $ &	$\times $ &	$\times $ &	\checkmark &	$\times $ &	$\times $ &	$\times $ &	$\times $ &	$\times $ \\
          \bottomrule
        
    \end{tabularx}
    \label{tab:suitability}
  \end{minipage}
  }
}
\end{table*}

\autoref{tab:suitability} presents frequently employed benchmark datasets for hallucination research. Certain datasets prioritize hallucination detection and analysis, while others help alleviate mitigation by introducing improved training and fine-tuning approaches. Another part of the benchmarks also targets implementing control mechanisms through factually improved grounding and low-confidence response filtering in real-time scenarios.

Benchmark analysis reveals a gap between current hallucination evaluation practices and the multi-stage manner in which hallucinations emerge across the LLM lifecycle. Existing datasets primarily focus on isolated tasks and output correctness while providing limited insight into upstream factors contributing to hallucination formation. Consequently, current benchmarks offer only partial diagnostic capability and often fail to capture aspects such as representation instability, reasoning inconsistency, and error propagation across lifecycle stages. Although task-specific datasets remain valuable, more comprehensive lifecycle-oriented benchmark frameworks are required to systematically evaluate hallucination behavior in LLMs.

\section{Discussions and Future Directions}
\label{sec:future_work}

This survey synthesizes existing research on hallucinations in LLMs through a lifecycle-oriented perspective that connects their causes, detection mechanisms, mitigation strategies, prevention techniques, and evaluation benchmarks. Addressing \textbf{RQ1}, the survey organizes the literature across the stages of data preparation, model training, and inference-time generation. The synthesis presented in \autoref{tab:rtm_causes}, together with the comparative analyses in \autoref{tab:mitigation_full} and \autoref{tab:prevention_full}, illustrates how weaknesses introduced in earlier stages of the model pipeline can propagate and manifest during generation.

Addressing \textbf{RQ2}, the analysis shows that hallucinations arise from interacting factors across the LLM lifecycle. Data-related issues such as biased distributions, incomplete coverage, and conflicting information introduce factual gaps during pretraining. Training-level factors including overfitting, misaligned optimization objectives, and architectural limitations can amplify these weaknesses, while inference-stage conditions such as decoding strategies, prompt ambiguity, and contextual limitations may trigger hallucinated responses.

The Detection framework (\autoref{fig:detection}) addresses \textbf{RQ3} by demonstrating that detection methods rely on observable signals produced at different lifecycle stages. Data-level approaches focus on dataset quality and external grounding, training-level techniques analyze internal model dynamics such as entropy and neuron activations, and inference-level methods rely on behavioral indicators such as response divergence, prompt sensitivity, and confidence estimation. However, most existing methods emphasize output-level verification rather than diagnosing internal mechanisms responsible for hallucinations.

The comparison of mitigation and prevention strategies addresses \textbf{RQ4} by showing that hallucination reduction requires coordinated interventions across the LLM lifecycle. Data-level approaches such as retrieval-augmented generation improve factual grounding, training-level methods including model editing and reinforcement learning refine internal representations, and inference-time techniques such as self-reflection and fact-checking act as runtime safeguards. Prevention strategies summarized in \autoref{tab:prevention_full} further highlight the importance of addressing hallucination risks earlier in the model development pipeline.

Addressing \textbf{RQ5}, the benchmark analysis reveals that many existing datasets focus primarily on output-level factual accuracy within narrow tasks and do not capture the lifecycle dynamics through which hallucinations emerge across data, training, and inference stages.

Collectively, these findings indicate that hallucinations in LLMs are not produced by a single isolated mechanism, but instead emerge through interacting vulnerabilities distributed across the data, training, and inference stages of the LLM lifecycle. The lifecycle-oriented synthesis further demonstrates that effective hallucination control requires coordinated strategies combining grounding, alignment, verification, intervention, and evaluation mechanisms operating across multiple stages of the model pipeline.

\noindent\textbf{Future Directions.} Future research should focus on improving the interpretability and reliability of LLMs through layer-aware diagnostic techniques capable of localizing hallucination-prone regions within transformer architectures. There is also a need for lifecycle-aware benchmarks that support cross-stage hallucination analysis, contextual grounding verification, multi-turn reasoning evaluation, and interpretability-aware diagnostics beyond output-level factual accuracy. Comparative studies of symbolic, statistical, and hybrid neurosymbolic approaches may further clarify effective hallucination control strategies, while adaptive grounding and verification mechanisms remain promising directions for improving reliability in high-stakes domains such as healthcare, law, and scientific research \cite{Lamba2025triggers,Lamba2025symloc}.

\section{Conclusion}

This survey provides a lifecycle perspective on hallucinations in LLMs by combining all that is known about why they happen, how to identify them, how to control them, how to prevent them, and how to measure them. By doing so, it presents a cohesive picture of how hallucinations happen in all aspects of the lifecycle of a LLMs. The key conclusion from this survey is that hallucinations are not a problem in a single step of a model’s lifecycle but are a reliability problem in general. By looking at all aspects of a model’s lifecycle from data collection to model training to inference, this lifecycle perspective highlights how all these reliability issues can combine to produce hallucinations in a model’s outputs. A comparative analysis of all that is known about how to control hallucinations and how to prevent them also highlights how a model’s reliability in terms of facts needs to be addressed in all aspects of a model’s lifecycle. The lifecycle perspective also highlights how current evaluations of a model’s performance using benchmarks are limited to only a part of a model’s lifecycle, specifically to its outputs. This again highlights a need for a more holistic evaluation framework to address hallucinations in a model’s outputs. Overall, this lifecycle perspective on understanding hallucinations in LLMs provides a practical foundation for tackling this problem in a more effective way. Further research in this area will depend on developing better tools to understand a model’s internal behavior, developing more robust benchmarks to test a model’s performance, and developing a holistic approach to controlling hallucinations in a model’s outputs by combining statistical learning with knowledge grounding.

\section*{Declarations}

\subsection*{Funding} 
This research did not receive any specific grant from funding agencies in the public, commercial, or not-for-profit sectors.

\subsection*{Conflict of interest}
The authors declare that they have no competing interests.

\subsection*{Consent for publication}
Not Applicable

\subsection*{Data availability} 
This study is a survey and does not involve the creation of new datasets. All datasets discussed in this paper are publicly available and cited appropriately.

\subsection*{Use of Generative AI}
The authors confirm that no generative AI tools were used to generate scientific content or conclusions in this manuscript. Any language assistance, if used, was limited to improving readability and did not affect the scientific substance of the work.

\bibliography{hallucination}% common bib file

\end{document}